\documentclass[runningheads]{llncs}
\usepackage{graphicx}
\usepackage{comment}
\usepackage{amsmath,amssymb}
\usepackage{color, colortbl}
\usepackage{siunitx}
\usepackage{booktabs}
\usepackage{multirow}
\usepackage{hyperref}
\usepackage{bm}
\usepackage{bbm}
\usepackage{caption}
\usepackage{float}
\usepackage{capt-of}
\usepackage[table,xcdraw,dvipsnames]{xcolor}
\usepackage{array}
\usepackage{pifont}

\newcommand{\bceloss}[2]{\sum_{i=1}^{V^2}\sum_{j=1}^{2} -\bigg[#2\text{log}~#1 + (1 - #2)\text{log}(1 - #1)\bigg]}
\newcommand{\tb}[1]{\textbf{#1}}
\newcommand{\code}[1]{\texttt{\small #1}}
\newcommand{\cmark}{\textcolor{green}{\ding{51}}}%
\newcommand{\xmark}{\textcolor{red}{\ding{55}}}%
\newcolumntype{P}[1]{>{\centering\arraybackslash}p{#1}}

\renewcommand{\footnotesize}{\scriptsize}

\definecolor{Gray}{gray}{0.9}

\begin{document}
\pagestyle{headings}
\mainmatter

\title{Occupancy Anticipation\\for Efficient Exploration and Navigation}
\titlerunning{Occupancy anticipation for efficient exploration and navigation}

\author{Santhosh K. Ramakrishnan\inst{1,2} \and
Ziad Al-Halah\inst{1} \and
Kristen Grauman\inst{1,2}}

\authorrunning{S. Ramakrishnan et al.}

\institute{The University of Texas at Austin, Austin TX 78712, USA \and
Facebook AI Research, Austin TX 78701, USA \\
\email{srama@cs.utexas.edu,~ziadlhlh@gmail.com,~grauman@cs.utexas.edu}\\
}

\vspace*{-2.0cm}
\begin{center}
    \footnotesize{In Proceedings of the European Conference on Computer Vision (ECCV), 2020}
\end{center}

\begingroup
\let\newpage\relax
\maketitle
\endgroup

\begin{abstract}
State-of-the-art navigation methods leverage a spatial memory to generalize to new environments, but their occupancy maps are limited to capturing the geometric structures directly observed by the agent. We propose \emph{occupancy anticipation}, where the agent uses its egocentric RGB-D observations to infer the occupancy state beyond the visible regions. In doing so, the agent builds its spatial awareness more rapidly, which facilitates efficient exploration and navigation in 3D environments.   By exploiting context in both the egocentric views and top-down maps our model successfully anticipates a broader map of the environment, with performance significantly better than strong baselines. Furthermore, when deployed for the sequential decision-making tasks of exploration and navigation, our model outperforms state-of-the-art methods on the Gibson and Matterport3D datasets. Our approach is the winning entry in the 2020 Habitat PointNav Challenge. 
\emph{Project page: \url{http://vision.cs.utexas.edu/projects/occupancy_anticipation/}}
\end{abstract}

\section{Introduction}

In visual navigation, an agent must move intelligently through a 3D environment in order to reach a goal. Visual navigation has seen substantial progress in the past few years, fueled by large-scale datasets and photo-realistic 3D environments~\cite{stanford2d3d,chang2017matterport,xia2018gibson,replica19arxiv},  simulators~\cite{xia2018gibson,ai2thor,mattersim,habitat19iccv}, and public benchmarks~\cite{das2018embodied,mattersim,habitat19iccv}.  Whereas traditionally navigation was attempted using purely geometric representations (i.e., SLAM), recent work shows the power of \emph{learned} approaches to navigation that integrate both geometry and semantics~\cite{zhu-iccv2017,gupta2017cognitive,savinov2018semi,mousavian2019visual,yang2018visual,chen2019learning}. Learned approaches operating directly on pixels and/or depth as input can be robust to noise~\cite{chen2019learning,chaplot2020learning} and can generalize well on unseen environments~\cite{gupta2017cognitive,habitat19iccv,yang2018visual,chaplot2020learning} \mbox{---even} outperforming pure SLAM given sufficient experience~\cite{habitat19iccv}.

One of the key factors for success in navigation has been the movement towards complex map-based architectures~\cite{gupta2017cognitive,parisotto2017neural,chen2019learning,chaplot2020learning} that capture both geometry~\cite{gupta2017cognitive,chen2019learning,chaplot2020learning} and semantics~\cite{gupta2017cognitive,parisotto2017neural,gordon2018iqa,henriques2018mapnet}, thereby facilitating efficient policy learning and planning. These learned maps allow an agent to exploit prior knowledge from training scenes when navigating in novel test environments.

Despite such progress, state-of-the-art approaches to navigation are limited to encoding \emph{what the agent actually sees in front of it}. In particular, they build maps of the environment  using only the \emph{observed} regions, whether via geometry~\cite{chen2019learning,henriques2018mapnet} or learning~\cite{gupta2017cognitive,parisotto2017neural,gordon2018iqa,chaplot2020learning}. Thus, while promising, today's models suffer from an important inefficiency:  to map a space in the 3D environment as free or occupied, the agent must directly see evidence thereof in its egocentric camera.

\begin{figure}[t]
    \centering
    \includegraphics[width=1.0\textwidth]{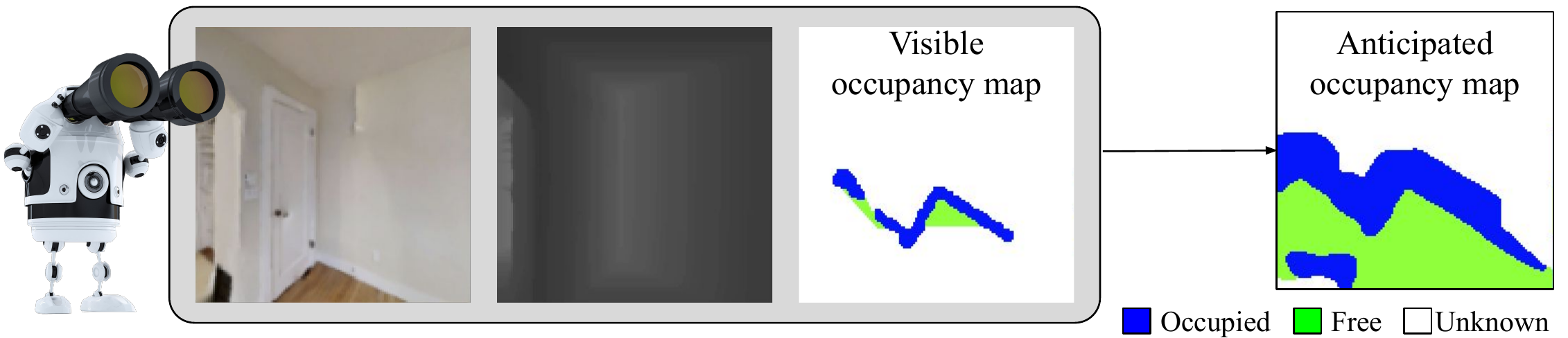}
    \caption{\small\textbf{Occupancy anticipation:} A robot's perception of the 3D world is limited by its field-of-view and obstacles (the visible map). We propose to anticipate occupancy for unseen regions (anticipated map) by exploiting the context from egocentric views.  We then train a deep reinforcement learning agent to move intelligently in a 3D environment, rewarding movements that improve the anticipated map.}
    \label{fig:intro_figure}
\end{figure}

Our key idea is to \emph{anticipate occupancy}. 
Rather than wait to directly observe a more distant or occluded region of the 3D environment to declare its occupancy status, the proposed agent infers occupancy for unseen regions based on the visual context in its egocentric views. For example, in Fig.~\ref{fig:intro_figure}, with only the partial observation of the scene, the agent could infer that it is quite likely that the wall extends to its right, a corridor is present on its left, and the region immediately in front of it is free space. Such intelligent extrapolation beyond the observed space would lead to more efficient exploration and navigation. To achieve this advantage, we introduce a model that anticipates occupancy maps from normal field-of-view RGB(D) observations, while aggregating its predictions over time in tight connection with learning a navigation policy. Furthermore, we incorporate the anticipation objective directly into the agent's exploration policy, encouraging movements in the 3D space that will efficiently yield broader and more accurate inferred occupancy maps.

We validate our approach on Gibson~\cite{xia2018gibson} and Matterport3D~\cite{chang2017matterport}, two 3D environment datasets spanning over 170 real-world spaces with a variety of obstacles and floor plans. Using only RGB(D) inputs to anticipate occupancy, the proposed agent learns to explore intelligently, achieving faster and more accurate maps compared to a state-of-the-art approach for neural SLAM~\cite{chaplot2020learning}, and navigating more efficiently than strong baselines.
Furthermore, for navigation under noisy actuation and sensing, our agent improves the state of the art, winning the 2020 Habitat PointNav Challenge~\cite{habitat-challenge} by a margin of 6.3 SPL points.

Our main contributions are: (1) a novel occupancy anticipation framework that leverages visual context from egocentric RGB(D) views; (2) a novel exploration approach that incorporates intelligent anticipation for efficient environment mapping, providing better maps in less time; and (3) successful navigation results that improve the state of the art.
\section{Related work}

\paragraph{Navigation} Classical approaches to visual navigation perform passive or active SLAM to reconstruct geometric point-clouds~\cite{thrun2002probabilistic,hartley2003multiple} or semantic maps~\cite{bao2012semantic,salas2013slam++}, facilitated by loop closures or learned odometry~\cite{cadena2016past,martinez2009bayesian,carrillo2012comparison}.  More recent work uses deep learning to learn navigation ~\cite{zhu-iccv2017,gupta2017cognitive,savinov2018semi,mousavian2019visual,yang2018visual,amodal,sax2018mid,shen2019situational} or exploration \cite{pathak2017curiosity,burada2018curiosity,savinov2018episodic,dinesh2018ltla,ramakrishnan2019emergence} policies in an end-to-end fashion. Explicit \emph{map-based} navigation models~\cite{gupta2017unifying,parisotto2017neural,gordon2018iqa,chen2019learning} usually outperform their implicit counterparts by being more sample-efficient, generalizing well to unseen environments, and even transferring from simulation to real robots~\cite{gupta2017cognitive,chaplot2020learning}. However, existing approaches only encode \emph{visible} regions for mapping (i.e., the ground plane projection of the observed or inferred depth). In contrast, our model goes beyond the visible cues and anticipates maps for unseen regions to accelerate navigation.

\paragraph{Layout estimation} Recent work predicts 3D Manhattan layouts of indoor scenes given 360 panoramas~\cite{zou2018layoutnet,yang2019dula,Sun_2019_CVPR,wu2019residential,Dhamo_2019_ICCV}. These methods predict structured outputs such as layout boundaries~\cite{zou2018layoutnet,Sun_2019_CVPR}, corners~\cite{zou2018layoutnet}, and floor/ceiling probability maps~\cite{yang2019dula}.  However, they do not extrapolate to unseen regions. FloorNet~\cite{liu2018floornet} and Floor-SP~\cite{cjc2019floorsp} use walkthroughs of previously scanned buildings to reconstruct detailed floorplans that may include predictions for the room type, doors, objects, etc. However, they assume that the layouts are polygonal, the scene is fully explored, and that detailed human annotations are available. Our occupancy map representation can be seen as a new way for the agent to infer the layout of its surroundings. Unlike any of the above approaches, our model does not make strict assumptions on the scene structure, nor does it require detailed semantic annotations. Furthermore, the proposed anticipation model is learned jointly with the exploration policy and without human guidance. Finally, unlike prior work, our goal is to accelerate navigation and map creation.

\paragraph{Scene completion} Past work in scene completion focuses on pixelwise reconstruction of 360 panoramas with limited glimpses~\cite{dinesh2018ltla,ramakrishnan2018sidekick,ramakrishnan2019emergence,seifi2019look}, inpainting~\cite{Pathak_2016_CVPR,iizuka2017globally,li2017generative}, and inferring unseen 3D structure and semantics~\cite{song2018im2pano3d,Yang_2019_CVPR}. While some methods allow pixelwise extrapolation outside the current field of view (FoV)~\cite{ramakrishnan2019emergence,song2018im2pano3d,Yang_2019_CVPR,jayaraman2018shapecodes}, they do not permit inferences about occluded regions in the scene. Our results show that this limitation is detrimental to successful occupancy estimation (cf.~our view extrapolation baseline).  SSCNet~\cite{song2016ssc} performs voxelwise geometric and semantic predictions for unseen 3D structures; however, it is computationally expensive, requires voxelwise semantic labels, limits predictions to the agent's FoV, and needs carefully curated viewpoints for training. In contrast, our approach predicts 2D occupancy from egocentric RGB(D) views, and it learns to do so in an active perception setting. Since the agent controls its own camera, the viewpoints tend to be more challenging than those in curated datasets of human-taken photos used in the scene completion literature~\cite{song2016ssc,song2018im2pano3d,dinesh2018ltla,ramakrishnan2018sidekick,Yang_2019_CVPR}.

\paragraph{Occupancy maps}
In robotics, methods for occupancy focus on building continuous representations of the world~\cite{o2012gaussian,ramos2016hilbert,senanayake2017deep}, mapping for autonomous driving~\cite{hoermann2018dynamic,mohajerin2019multi,sless2019self,lu2019hallucinating,muller2018driving}, and indoor robot navigation~\cite{katyal2019uncertainty,elhafsi2019map,shrestha2019learned}. Prior extrapolation methods assume wide FoV LIDAR inputs, only exploit geometric cues from top-down views, and demonstrate results in relatively simple 2D floorplans devoid of non-wall obstacles~\cite{katyal2018occupancy,katyal2019uncertainty,elhafsi2019map,shrestha2019learned}. In contrast, our approach does not require expensive LIDAR sensors. It operates with standard RGB(D) camera inputs, and it exploits both semantic and geometric context from those egocentric views to perform accurate occupancy anticipation. Furthermore, we demonstrate efficient navigation in visually rich 3D environments with challenging obstacles other than walls. Finally, unlike prior work, our anticipation models are learned jointly with a navigation policy that rewards accurate anticipatory mapping.
\section{Approach}

We propose an occupancy anticipation approach for efficient exploration and navigation. Our model anticipates areas not directly visible to the agent because of occlusion (e.g., behind a table, around a corner) or due to being outside its FoV. The agent's first-person view is provided in the form of RGB-D images (see Fig.~\ref{fig:approach_figure} left). The goal is to anticipate the occupancy for a fixed region in front of the agent, and integrate those predictions over time as the agent moves about.

Next, we define the task setup and notation, followed by our approach for occupancy anticipation (Sec.~\ref{sec:occupancy_model}) and a new formulation for exploration that rewards correctly anticipated regions (Sec.~\ref{sec:anticipation_reward}). Then, we explain how our occupancy anticipation model can be integrated into a state-of-the-art approach~\cite{chaplot2020learning} for autonomous  exploration and navigation in 3D environments (Sec.~\ref{sec:navigation_model}).

\subsection{Occupancy anticipation model}\label{sec:occupancy_model}

\begin{figure}[t]
    \centering
    \includegraphics[width=1.0\textwidth]{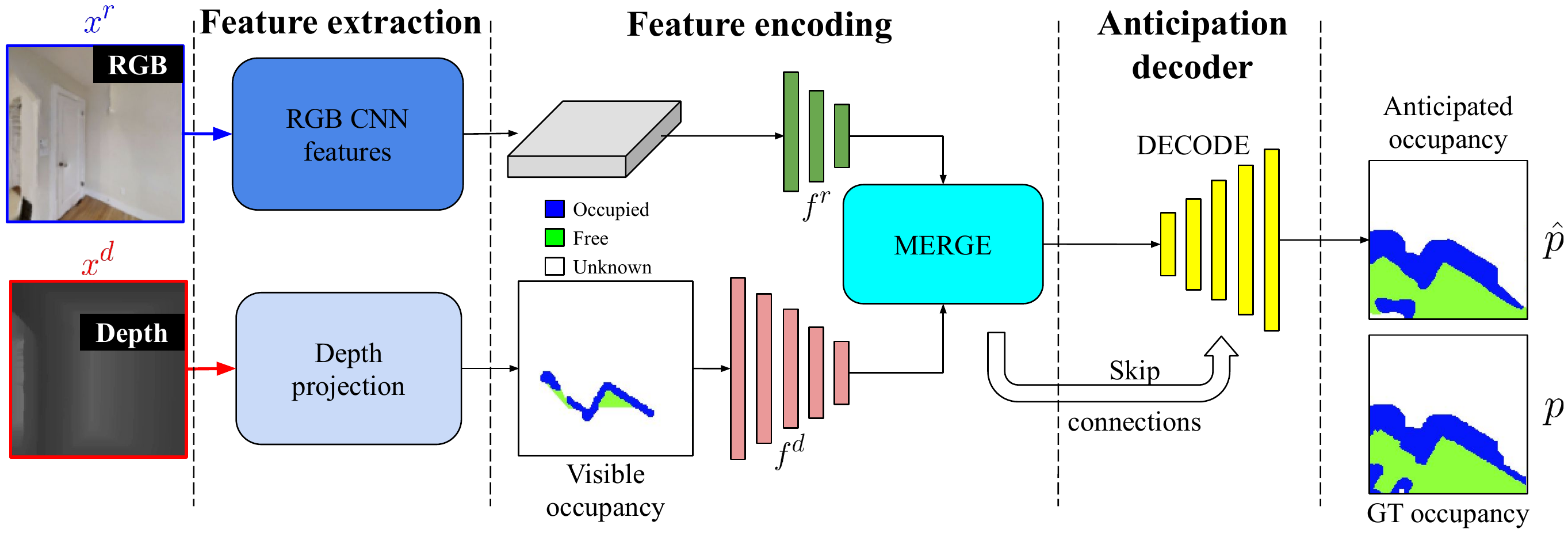}
    \caption{\small Our occupancy anticipation model uses RGB(D) inputs to extract features, and processes them using a UNet to anticipate the occupancy. The depth map is projected to the ground plane to obtain the preliminary visible occupancy map.  See text.}  
    \label{fig:approach_figure}
\end{figure}

We formulate occupancy anticipation as a pixelwise classification task. The egocentric occupancy is represented as a two-channel top-down map $p \in [0,1]^{2\times V \times V}$ which comprises a local area of $V \times V$ cells in front of the camera. Each cell in the map represents a $25\si{mm} \times 25\si{mm}$ region. The two channels contain the probabilities (confidence values) of the cell being occupied and explored, respectively. A cell is considered to be occupied if there is an obstacle, and it is explored if we know whether it is occupied or free. For training, we use the 3D meshes of indoor environments (Sec.~\ref{sec:exp_setup}) to obtain the ground-truth local occupancy of a $V \times V$ region in front of the camera, which includes parts that may be occluded or outside the field of view (Fig.~\ref{fig:approach_figure}, bottom right). 

Our occupancy anticipation model consists of three main components (Fig.~\ref{fig:approach_figure}): \\ 
\noindent\textbf{(1) Feature extraction:} Given egocentric RGB-D inputs, we compute: 

\noindent\emph{RGB CNN features:} We encode the RGB images using blocks 1 and 2 of a ResNet-18 that is pre-trained on ImageNet, followed by three additional convolution layers that prepare these features to be passed forward with the visible occupancy map. This step extracts a mixture of textural and semantic features.

\noindent\emph{Depth projection:} We estimate a map of occupied, free, and unknown space by setting height thresholds on the point cloud obtained from depth and camera intrinsics~\cite{chen2019learning}. Consistent with past work~\cite{chen2019learning,chaplot2020learning}, we restrict the projection-based estimates to points within $\sim3\si{m}$, the range at which modern depth sensors would provide reliable results. This yields the initial visible occupancy map.\\

\noindent\textbf{(2) Feature encoding:} Given the RGB-D features, we independently encode them using UNet~\cite{ronneberger2015u} encoders and project them to a common feature space. We encode the depth projection features using a stack of five convolutional blocks which results in features $\bm{f}^{d} = f_{1:5}^{d}$. Since the RGB features are already at a lower resolution, we use only three convolutional blocks to encode them, which results in features $\bm{f}^{r} = f_{3:5}^{r}$. We then combine these features using the \textsc{Merge} module which contains layer-specific convolution blocks to merge each $[\bm{f}^{r}_{i}, \bm{f}^{d}_{i}]$: 

\begin{equation}
\bm{f} = \textsc{merge}(\bm{f}^{d}, \bm{f}^{r}).
\end{equation}
For experiments with only the depth modality, we skip the RGB feature extractor and $\textsc{Merge}$ layer and directly use the occupancy features obtained from the depth image. For experiments with only the RGB modality, we learn a model to infer the visible occupancy features from RGB (to be defined at the end of Sec.~\ref{sec:exp_setup}) and use that instead of the features computed from the depth image. \\

\noindent\textbf{(3) Anticipation decoding:}
Given the encoded features $\bm{f}$, we use a UNet decoder that outputs a $2 \times V \times V$ tensor of probabilities:
\begin{equation}
    \hat{p} = \sigma(\textsc{Decode}(\bm{f})),
\end{equation} 
where $\hat{p} \in [0, 1]^{2\times V\times V}$ is the estimated egocentric occupancy and $\sigma$ is the sigmoid activation function. For training the occupancy anticipation model, we use binary cross entropy loss per pixel and per channel:
\begin{equation}
\label{eqn:loss_anticp}
L = \bceloss{\hat{p}_{ij}}{p_{ij}},
\end{equation} 
where $p$ is the ground-truth (GT) occupancy map that is derived from the 3D mesh of training environments (see Sec.~S5 in Supp. for details).  

So far, we have presented our occupancy anticipation approach supposing a single RGB-D observation as input. However, our model is ultimately used in the context of an embodied agent that moves in the environment and actively collects a sequence of RGB-D views to build a complete map of the environment. Next, we introduce a new reward function that utilizes the agent's anticipation performance to guide its exploration during training.

\subsection{Anticipation reward for exploration policy learning}
\label{sec:anticipation_reward}

In \emph{visual exploration}, an agent must quickly map a new environment without having a specified target. Prior work on exploration~\cite{chen2019learning,fang2019scene,chaplot2020learning,ramakrishnan2020exploration} often uses area-coverage---the area seen in the environment during navigation---as a reward function to guide exploration.  However, the traditional area-coverage approach is limited to rewarding the agent only for \emph{directly seeing} areas.  Arguably, an ideal exploration agent would obtain an accurate and complete map of the environment \emph{without} necessarily directly observing all areas.

Thus, we propose to encourage exploratory behaviors that yield a correctly \emph{anticipated} map. In this case, the occupancy entries in the map need not be obtained via direct agent observations to register a reward; it is sufficient to correctly infer them. In particular, we reward agent actions that yield accurate occupancy predictions for the global environment map, i.e., the number of grid cells where the predicted occupancy matches the layout of the environment.

More concretely, let $\hat{m}_{t} \in [0, 1]^{2 \times G \times G}$ be the global environment map obtained by anticipating occupancy for the RGB-D observations $\{x^{r}_{1:t}, x^{d}_{1:t}\}$ from time $1$ to $t$, and then geometrically registering the predictions to a single global map based on the agent's pose estimates at each time step (see Fig.~\ref{fig:exploration_model}). Note $G > V$. Let $m$ be the ground-truth layout of the environment. Then, the unnormalized accuracy of a map prediction $\hat{m}$ is measured as follows: 
\begin{equation}
\label{eqn:anticipation_accuracy}
   \text{Accuracy}(\hat{m}, m) = \sum_{i=1}^{G^2} \sum_{j=1}^{2}  \mathbbm{1}[\hat{m}_{ij} = m_{ij}],
\end{equation}  
where $\mathbbm{1}[\hat{m}_{ij} = m_{ij}]$ is an indicator function that returns one if $\hat{m}_{ij} = m_{ij}$ and zero otherwise. We reward the increase in map accuracy from time $t-1$ to $t$:
\begin{equation}
\label{eqn:anticipation_reward}
    R_{t}^{anticp} = \text{Accuracy}(\hat{m}_{t}, m) - \text{Accuracy}(\hat{m}_{t-1}, m).
\end{equation}

This function rewards actions leading to correct global map predictions, irrespective of whether the agent actually \emph{observed} those locations. For example, if the agent correctly anticipates free space behind a table and is rewarded for that, it then learns to avoid spending additional time around tables in the future to observe that space directly. Resources can be instead allocated to visiting more interesting regions that are harder to anticipate. Additionally, this reward provides a better learning signal while training under noisy conditions by accounting for mapping errors arising from noisy pose and map predictions. Thus, our approach encourages more intelligent exploration behavior by injecting our anticipated occupancy idea directly into the agent's sequential decision-making.

\subsection{Exploration and navigation with occupancy anticipation}
\label{sec:navigation_model}

\begin{figure}[t]
    \centering
    \includegraphics[width=1.0\textwidth,trim=0 7.5cm 0 0,clip]{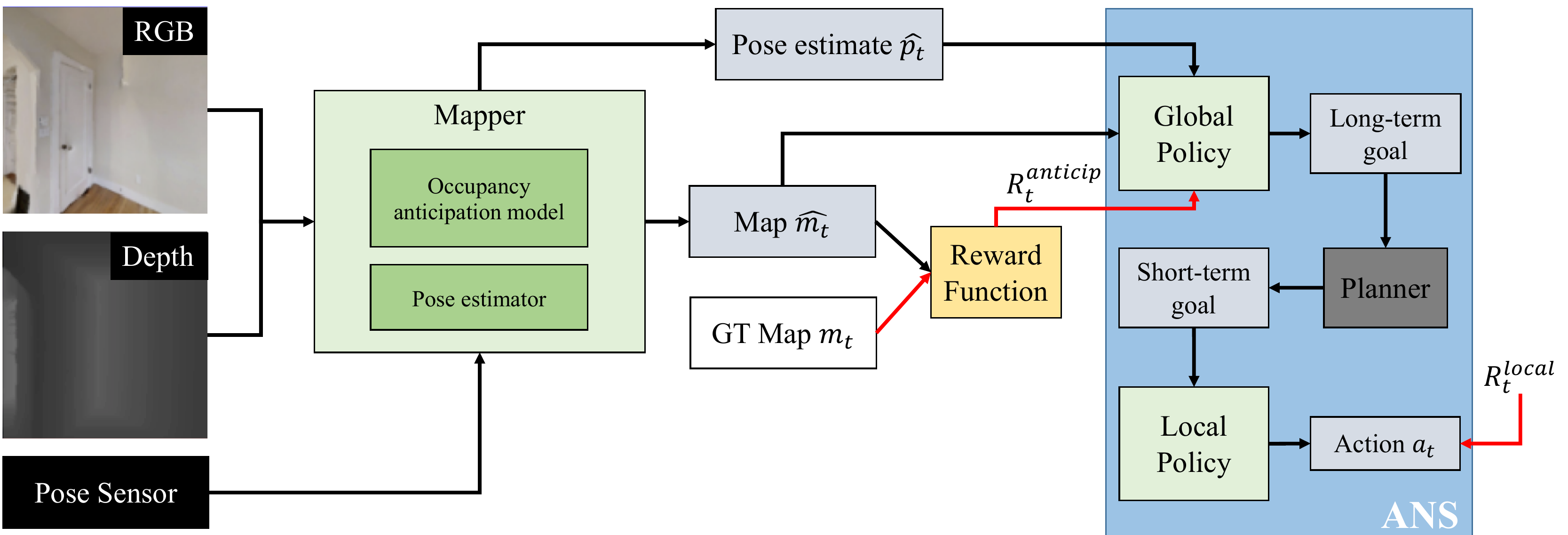}
    \caption{\small\tb{Exploration with occupancy anticipation:} We introduce two key upgrades to the original Active Neural SLAM (ANS) model~\cite{chaplot2020learning} (see text): (1) We replace the projection unit in the mapper with our occupancy anticipation model (see Fig.~\ref{fig:approach_figure}). (2) We replace the area-coverage reward function with the proposed reward (Eqn.~\ref{eqn:anticipation_reward}), which encourages the agent to efficiently explore and build accurate maps through occupancy anticipation. Note that the reward signals (in red) are provided only during training.}
    \label{fig:exploration_model}
\end{figure}

Having defined the core occupancy anticipation components, we now demonstrate how our model can be used to benefit embodied navigation in 3D environments. We consider both exploration (discussed above) and \emph{PointGoal navigation}~\cite{savva2017minos,anderson2018evaluation}, a.k.a PointNav, where the agent must navigate efficiently to a target specified by a displacement vector from the agent's starting position.

For both tasks, we adapt the state-of-the-art Active Neural SLAM (ANS) architecture~\cite{chaplot2020learning} that previously achieved the best exploration results in the literature and was the winner of the 2019 Habitat PointNav challenge. However, our anticipation model is generic and can be easily integrated with most map-based embodied navigation models~\cite{gupta2017cognitive,chen2019learning,gan2019look}. 

The ANS model is a hierarchical, modular policy for exploration that consists of a mapper, a planner, a local policy, and a global policy (shown in Fig.~\ref{fig:exploration_model}). Given RGB images, the mapper estimates the egocentric occupancy and agent pose, and then temporally aggregates the maps into a global top-down map using the pose estimates. At regular time intervals $\Delta$, the global policy picks a location on the global map to explore. A shortest-path planner decides what trajectory to take from the current position to the target and picks an intermediate goal (within $1.25\si{m}$) to navigate to. The local policy then selects actions that lead to the intermediate goal; it gets another intermediate goal upon reaching the current goal. See~\cite{chaplot2020learning} for details.  Critically, and like other prior work, the model of~\cite{chaplot2020learning} is supervised to generate occupancy estimates based solely on the \emph{visible} occupancy obtained from the egocentric views.

We adapt ANS by modifying the mapper and the reward function. For the mapper, we replace the projection unit from ANS with our anticipation model (see Fig.~\ref{fig:exploration_model}). Additionally, we account for incorrect occupancy estimates in two ways: (1) we filter out high entropy predictions and (2) we maintain a moving average estimate of occupancy at each location in the global map (see Sec.~S7 in Supp.). For the reward function, we use the anticipation-based reward presented in Sec.~\ref{sec:anticipation_reward}.

We train the exploration policy with our anticipation model end-to-end, as this allows adapting to the changing distribution of the agent's inputs. Both the local and the global reinforcement learning policies are trained with Proximal Policy Optimization (PPO)~\cite{schulman2017proximal}. In our model, the reward of the global policy is our anticipation-based reward defined in Eqn.~\ref{eqn:anticipation_reward}.  This replaces the traditional area-coverage reward used in ANS and other current models~\cite{chen2019learning,chaplot2020learning,ramakrishnan2020exploration}, which rewards the increment in the actual area seen, not the correctly registered area in the map. The reward for the local policy is simply based on the reduction in the distance to the local goal:  $R_{t}^{local} = d_{t-1} - d_{t}$,  where $d$ is the Euclidean distance between the current position and the local goal. 
\section{Experiments}

In the following experiments we demonstrate that 1) our occupancy anticipation module can successfully infer unseen parts of the map (Sec.~\ref{sec:exp_occupancy}) and 2) trained together with an exploration and navigation policy, it accelerates active mapping and navigation in new environments (Sec.~\ref{sec:exp_exploration} and Sec.~\ref{sec:exp_navigation}).

\subsection{Experimental setup}
\label{sec:exp_setup}

We use the Habitat~\cite{habitat19iccv} simulator along with Gibson~\cite{xia2018gibson} and Matterport3D~\cite{chang2017matterport} environments. Each dataset contains around $90$ challenging large-scale photo-realistic 3D indoor environments such as houses and office buildings.  On average, the Matterport3D environments are larger. Our observation space consists of $128 \times 128$ RGB-D observations and odometry sensor readings that denote the change in the agent's pose $x, y, \theta$. Our action space consists of three actions: \textsc{move-forward} by $25\si{cm}$, \textsc{turn-left} by $10^{\circ}$, \textsc{turn-right} by $10^{\circ}$. For navigation, we add a \textsc{stop} action, which the agent emits when it believes it has reached the goal. We simulate noisy actuation and odometer readings for realistic evaluation (see Sec.~S6 in Supp.).

We train our exploration models on Gibson, and then transfer them to PointGoal navigation on Gibson and exploration on Matterport3D. We use the default train/val/test splits provided for both datasets~\cite{habitat19iccv} with disjoint environments across the splits.  For evaluation on Gibson, we divide the validation environments into small (area less than $36\si{m^2}$) and large (area greater than $36\si{m^2}$) to observe the influence of environment size on results. For policy learning, we use the Adam optimizer and train on episodes of length $1000$ for $1.5-2$ million frames of experience. Please see Sec.~S8 in Supp.~for more details. \\

\noindent\textbf{Baselines:} We define baselines based on prior work:
\begin{itemize}
    \item \textbf{ANS(rgb)}~\cite{chaplot2020learning}: This is the state-of-the-art Active Neural SLAM approach for exploration and navigation. We use the original mapper architecture~\cite{chaplot2020learning}, which infers the visible occupancy from RGB.\footnote{We use our own implementation of ANS since authors' code was unavailable at the time of our experiments. See Sec.~S7 in Supp.~for details.}
    \item \textbf{ANS(depth)}: We use depth projection to infer the visible occupancy (similar to~\cite{chen2019learning}) instead of predicting it from RGB.
    \item \textbf{View-extrap.}: We extrapolate an $180^{\circ}$ FoV depth map from $90^{\circ}$ FoV RGB-D and project it to the top-down view.  This is representative of scene completion approaches~\cite{song2018im2pano3d,Yang_2019_CVPR}. See Sec.~S11 in Supp.~for network details.
    \item \textbf{OccAnt(GT)}: This is an upper bound that cheats by using the ground-truth anticipation maps for exploration and navigation.
\end{itemize}

We implement all baselines on top of the ANS framework. Our goal is to show the impact of our occupancy model, while fixing the backbone navigation architecture and policy learning approach across methods for a fair comparison. We consider three versions of our models based on the input modality:

\begin{itemize}
    \item \textbf{OccAnt(depth)}: anticipate occupancy given the visible occupancy map. 
    \item \textbf{OccAnt(rgb)}: anticipate occupancy given only the RGB image. We replace the depth projections in Fig.~\ref{fig:approach_figure} with the pre-trained ANS(rgb) estimates (kept frozen throughout training).
    \item \textbf{OccAnt(rgbd)}: anticipate occupancy given the full RGB-D inputs.
\end{itemize}
By default, our methods use the proposed anticipation reward from Sec.~\ref{sec:anticipation_reward}. We denote ablations without this reward as ``w/o AR". \\

\begin{table}[!t]
\begin{minipage}{\linewidth}
\centering
\scalebox{0.8}{
\begin{tabular}{@{}p{4cm}p{1.1cm}p{1.1cm}p{1.1cm}p{1.1cm}p{1.1cm}p{1.1cm}@{}}
\toprule
\multirow{2}{*}{Method}   & \multicolumn{3}{c}{IoU \%}          & \multicolumn{3}{c}{F1 score \%}   \\ \cmidrule(r){2-4} \cmidrule(r){5-7}
                          & free      & occ.      & mean        & free      & occ.      & mean      \\ \midrule
 all-free                 & 30.1      & 0         & 15.1        & 43.6      & 0         & 21.8      \\
 all-occupied             & 0         & 25.1      & 12.6        & 0         & 39.2      & 19.6      \\
 ANS(rgb)                 & 12.1      & 14.9      & 13.5        & 19.6      & 24.9      & 22.5      \\
 ANS(depth)               & 14.5      & 24.1      & 19.3        & 23.1      & 37.6      & 30.4      \\
 View-extrap.             & 15.5      & 26.4      & 21.0        & 25.0      & 40.4      & 32.7      \\ \midrule
 OccAnt(rgb)              & 44.4      & 47.9      & 46.1        & 58.2      & 62.9      & 60.6      \\
 OccAnt(depth)            & 50.4      & \tb{61.9} & 56.1        & 63.8      & \tb{75.0} & 69.4      \\
 OccAnt(rgbd)             & \tb{51.5} & 61.5      & \tb{56.5}   & \tb{64.9} & 74.8      & \tb{69.8} \\
 \bottomrule
\end{tabular}
}
\caption{\small \textbf{Occupancy anticipation results} on the Gibson validation set. Our models, OccAnt($\cdot$), substantially improve the map quality and extent, showing the advantage of learning to anticipate 3D structures beyond those directly observed.}
\label{tab:occ_anticipation}
\end{minipage}
\end{table}

\subsection{Occupancy anticipation results}\label{sec:exp_occupancy}

First we evaluate the per-frame prediction accuracy of the mapping models trained during exploration.  We evaluate on a separate dataset of images sampled from validation environments in Gibson at uniform viewpoints from discrete locations on a $1\si{m}$ grid, a total of $1,034$ (input, output) samples. This allows standardized evaluation of the mapper, independent of the exploration policy. 

To quantify the local occupancy maps' accuracy, we compare the predicted maps to the ground truth. We report the Intersection over Union (IoU) and F1 scores for the ``free" and ``occupied" classes independently. In addition to the baselines from Sec.~\ref{sec:exp_setup}, we add two naive baselines that classify all locations as free (all-free), or occupied (all-occupied). 

Table~\ref{tab:occ_anticipation} shows the results.  Our anticipation models OccAnt are substantially better than all the baselines.  Comparing different modalities, OccAnt(depth) is much better than OccAnt(rgb) under all the metrics. This makes sense, as visible occupancy is directly computable from the depth input, but must be inferred for RGB (see Fig.~\ref{fig:occupancy_anticipation_qualitative}). Interestingly, the rgbd models are not better than the depth-only models, likely because (1) geometric cues are more easily learned from depth than RGB, and (2) the RGB encoder contains significantly more parameters and could lead to overfitting.  See Table~S5 in Supp.~for network sizes. Overall, Table~\ref{tab:occ_anticipation} demonstrates our occupancy anticipation models successfully broaden the coverage of the map beyond the visible regions.

\subsection{Exploration results}\label{sec:exp_exploration}

Next we deploy our models for visual exploration. The agent is given a limited time budget ($T$=1000) to intelligently explore and build a 2D top-down occupancy map of a previously unseen environment. 

To quantify exploration, we measure both map quality and speed (number of agent actions):
(1) \textbf{Map accuracy (\si{m^2}):} the area in the global map built during exploration (both free and occupied) that matches with the ground-truth layout of the environment. The map is built using predicted occupancy maps which are registered using estimated pose (may be noisy). Note that this is an unnormalized accuracy measure (see Eqn.~\ref{eqn:anticipation_accuracy}). (2) \textbf{IoU:} the intersection over union between that same global map and the ground-truth layout of the environment. (3) \textbf{Area seen (\si{m^2}):} the amount of free and occupied regions \emph{directly seen} during exploration. The map for this metric is built using ground-truth pose and depth-projections (similar to~\cite{chen2019learning,chaplot2020learning}). (4) \textbf{Episode steps:} the number of actions taken by the agent. While the first two metrics measure the quality of the created map, the latter two are a function of how (and how long) the agent moved to get that map. Higher accuracy in fewer steps or lower area-seen is better.\\

All agents are trained on 72 scenes from Gibson under noisy odometry and actuation (see Sec.~\ref{sec:exp_setup}), and evaluated on Gibson and Matterport3D under both noisy and noise-free conditions. 

\begin{table}[H]
\begin{minipage}{\linewidth}
    \centering
    \includegraphics[width=1.0\textwidth]{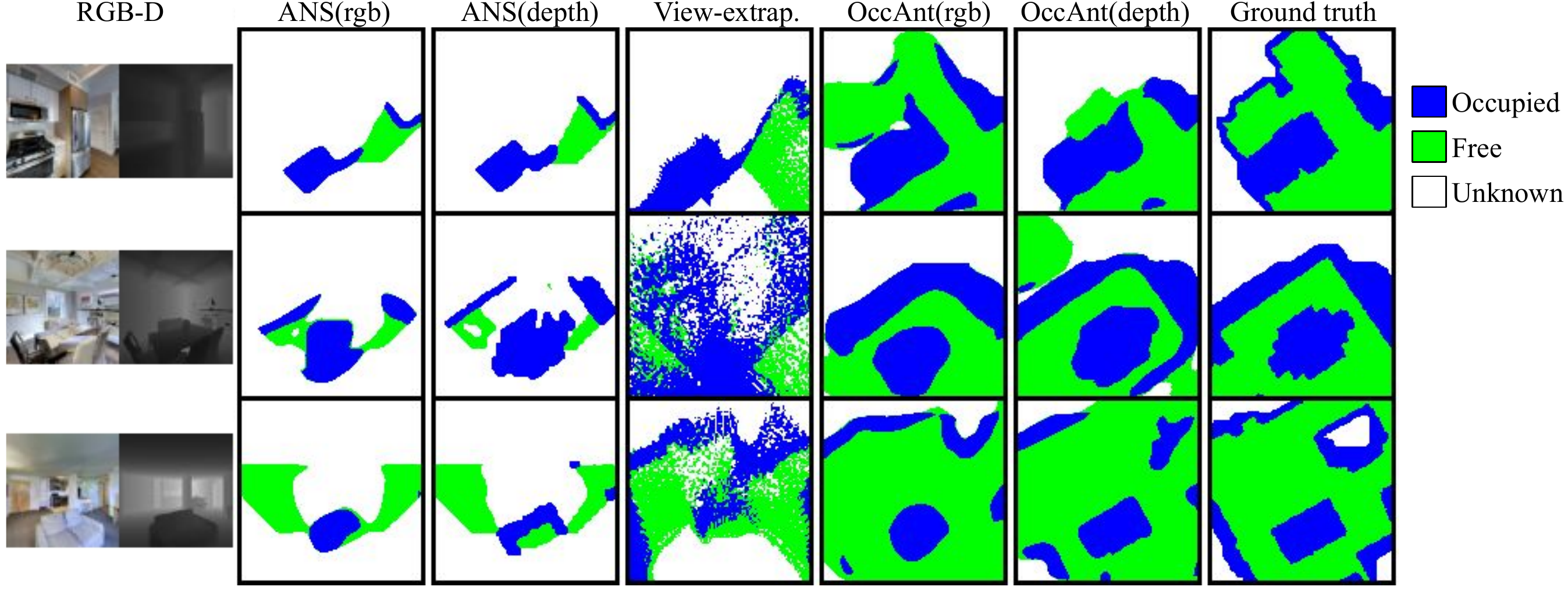}
    \captionof{figure}{\small \textbf{Per-frame local occupancy predictions:} First and last columns show the RGB-D input and anticipation ground-truth, respectively. ANS(*) are restricted to only predicting occupancy for visible regions. View-extrap. extrapolates, but is unable to predict occupancy for occluded regions (first row) and struggles to make correct predictions in cluttered scenes (second row). Our model successfully anticipates with either RGB or depth. For example, in the first row, we successfully predict the presence of a corridor and another room on the left. In the second row, we successfully predict the presence of navigable space behind the table. In the third row, we are able to correctly anticipate the free space behind the chair and the corridor to the right.}
    \label{fig:occupancy_anticipation_qualitative}
\end{minipage}
\begin{minipage}{\linewidth}
    \centering
    \includegraphics[width=0.98\textwidth]{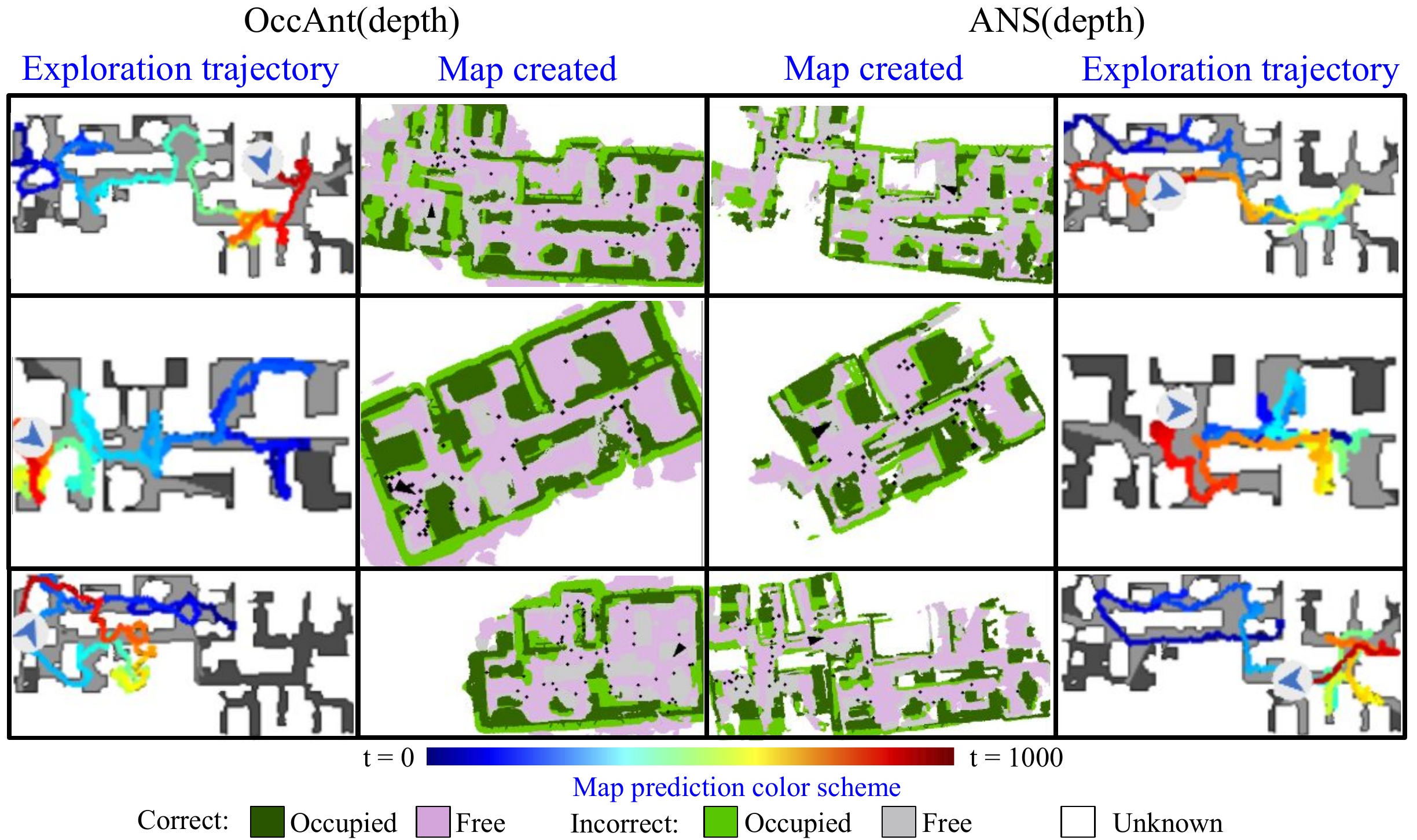}
    \captionof{figure}{\small\textbf{Exploration examples:} We compare OccAnt with ANS~\cite{chaplot2020learning} in Gibson under noisy actuation and odometry. The exploration trajectories and the corresponding maps are shown at the extremes and center, respectively. \textbf{Row 1:} Both methods cover similar area, but our method better anticipates the unseen parts with fewer registration errors. \textbf{Row 2:} Our method achieves better area coverage and mapping quality whereas the baseline gets stuck in a small room for extended periods of time. \textbf{Row 3:} A failure case for our method, where it gets stuck in one part of the house after anticipating that a narrow corridor leading to a different room was occupied.} 
    \label{fig:qual}
\end{minipage}
\end{table}

\begin{table}[!t]
\begin{minipage}{\linewidth}
    \centering
    \includegraphics[width=1.00\textwidth,trim=0 1.5cm 6.4cm 0]{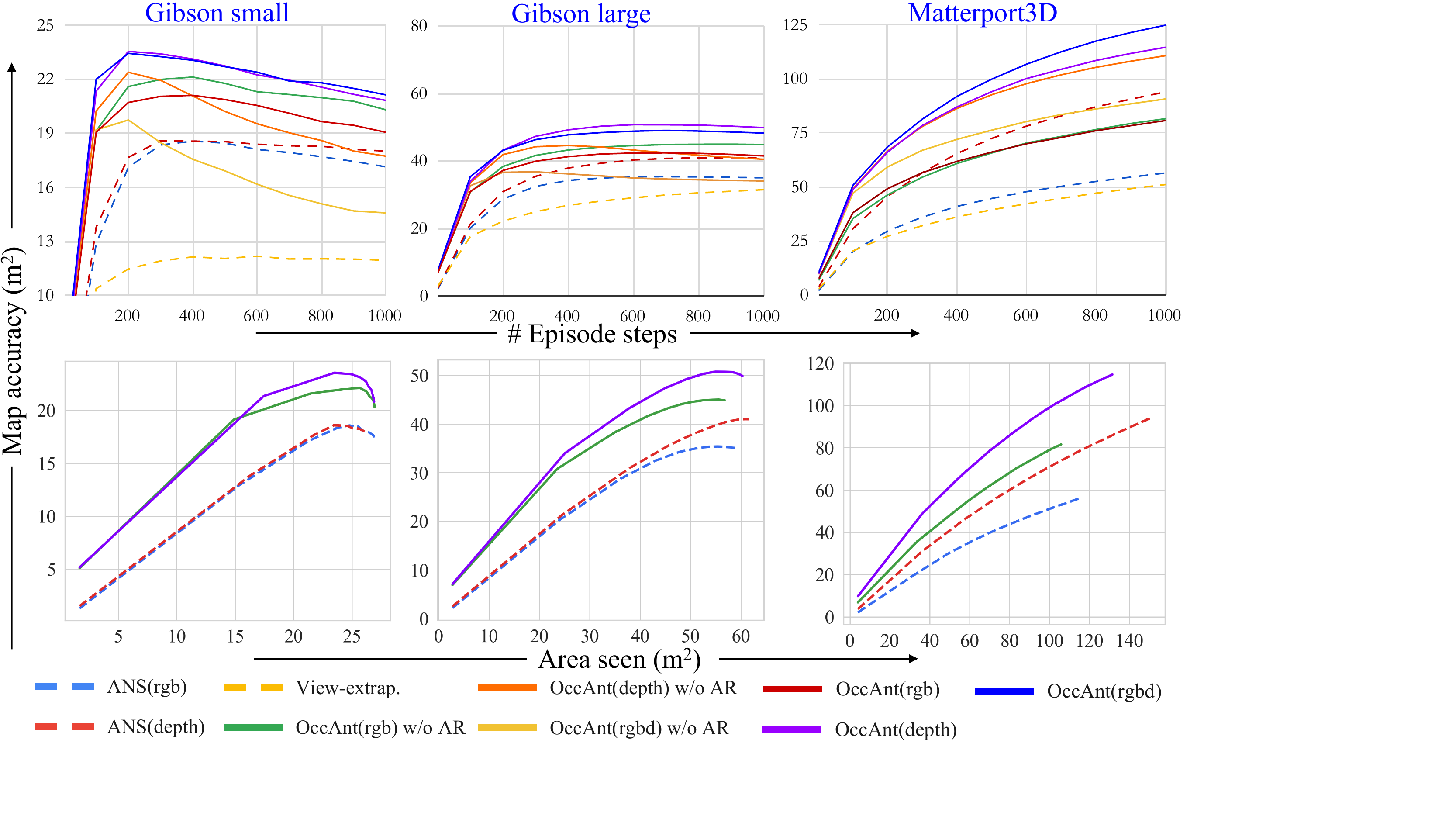}
    \captionof{figure}{\small\textbf{Exploration results:} Map accuracy (\si{m^2}) as a function of episode duration (top row) and area seen (bottom row) for Gibson (small and large splits) and Matterport3D under noisy conditions (see Sec.~S1 in Supp.~for noise-free). Higher and steeper curves are better.  \textbf{Top:} Our OccAnt approach rapidly attains higher map accuracy than the baselines (dotted lines). \textbf{Bottom:} OccAnt achieves higher map accuracy for the same area seen (we show the best variants here to avoid clutter). These results show the agent \emph{actively moves better} to explore the environment with occupancy anticipation.}
    \label{fig:exploration_results}
\end{minipage}

\begin{minipage}{\linewidth}
    \centering
    \scalebox{0.77}{
    \begin{tabular}{@{}lcccccccccccc@{}}
    \toprule
    \rowcolor{Gray}                &                                   \multicolumn{6}{c}{Noisy test conditions}                                  &                                 \multicolumn{6}{c}{Noise-free test conditions}                              \\ \midrule
                                   & \multicolumn{2}{c}{Gibson small}    & \multicolumn{2}{c}{Gibson large}  & \multicolumn{2}{c}{Matterport3D}   & \multicolumn{2}{c}{Gibson small}    & \multicolumn{2}{c}{Gibson large}  & \multicolumn{2}{c}{Matterport3D}  \\ \cmidrule(r){2-3} \cmidrule(r){4-5} \cmidrule(r){6-7}\cmidrule(r){8-9} \cmidrule(r){10-11} \cmidrule(r){12-13}
    Method                         &      Map acc.      &       IoU      &      Map acc.   &     IoU         &      Map acc.    &      IoU        &      Map acc.     &       IoU       &      Map acc.     &      IoU      &      Map acc.    &       IoU      \\ \midrule
    ANS(rgb)~\cite{chaplot2020learning}& 18.5           &      55        &     35.0        &     47          &     44.7         &       18        &       22.4        &       76        &       43.4        &       64      &     53.4         &       23       \\
    ANS(depth)                     &     18.5           &      56        &     39.4        &     53          &     72.5         &       26        &       21.4        &       74        &       48.0        &       72      &     85.9         &       34       \\
    View-extrap.                   &     12.0           &      26        &     28.1        &     27          &     39.4         &       14        &       12.1        &       27        &       26.5        &       27      &     33.9         &       13       \\\midrule
    OccAnt(rgb) w/o AR             &     21.8           &      66        &     44.2        &     57          &     65.8         &       23        &       22.6        &       71        &       45.2        &       60      &     64.4         &       24       \\
    OccAnt(depth) w/o AR           &     20.2           &      58        &     44.2        &     54          &     92.7         &       29        &   \tb{24.9}       &   \tb{84}       &   \tb{54.1}       &   \tb{75}     &.  \tb{104.7}     &   \tb{38}.     \\
    OccAnt(rgbd) w/o AR            &     16.9           &      45        &     35.6        &     40          &     76.3         &       23        &       24.8        &   \tb{84}       &       52.0        &       71      &     98.7         &       34       \\
    OccAnt(rgb)                    &     20.9           &      62        &     42.1        &     54          &     66.2         &       22        &       22.3        &       70        &       43.5        &       58      &     64.4         &       22       \\
    OccAnt(depth)                  & \tb{22.7}          &  \tb{71}       & \tb{50.3}       & \tb{67}         &     94.1         &   \tb{33}       &       24.8        &       83        &       53.1        &       74      &     96.5         &       35       \\ 
    OccAnt(rgbd)                   & \tb{22.7}          &  \tb{71}       &     48.4        &     62          & \tb{99.9}        &       32        &       24.5        &       82        &       51.0        &       69      &    100.3         &       34       \\ \midrule
    \rowcolor{Gray} OccAnt(GT)     &     21.7           &      67        &     51.9        &     63          &       -          &         -       &       26.1        &       93        &       65.4        &       91      &       -          &       -        \\ \bottomrule
    \end{tabular}
    }
\caption{\small\textbf{Timed exploration results:} Map quality at $T$=500 for all models and datasets. See text for details.}
\label{tab:exploration_results}
\end{minipage}\hfill
\end{table}

Fig.~\ref{fig:exploration_results} shows the exploration results. Our approach generally outperforms the baselines, improving the map quality more rapidly, whether in terms of time (top row) or area seen (bottom row). When compared on a same-modality basis, we see that OccAnt(rgb) converges much faster than ANS(rgb). Similarly, OccAnt(depth) is able to rapidly improve the map quality and outperforms ANS(depth) on all cases. This apples-to-apples comparison shows that anticipating occupancy leads to much more efficient mapping in unseen environments. Again, using depth generally provides more reliable mapping than pure RGB.

Furthermore, the proposed anticipation reward generally provides significant benefits to map accuracy in the noisy setting (compare our full model to the ``w/o AR" models in Fig.~\ref{fig:exploration_results}). While map accuracy generally increases over time for noise-free conditions (see Sec.~S1 in Supp.), it sometimes saturates early or even declines slightly over time in the noisy setting as noisy pose estimates accumulate and hurt map registration accuracy.  This is most visible in Gibson small (top left plot). However, our anticipatory reward alleviates this decline.

Table~\ref{tab:exploration_results} summarizes the map accuracy and IoU for all methods at $T$=500. Our method obtains significant improvements, supporting our claim that occupancy anticipation accelerates exploration and mapping. Additionally, perfect anticipation with the OccAnt(GT) model gives comparably good noisy exploration,  and good gains in noise-free exploration (+10-20$\%$  IoU). This shows that there is indeed a lot of mileage in anticipating occupancy; our model moves the state-of-the-art towards this ceiling. Fig.~\ref{fig:qual} shows example exploration trajectories and the final global map predictions on Gibson.

\subsection{Navigation results}\label{sec:exp_navigation}

Next we evaluate the utility of occupancy anticipation for quickly reaching a target. In PointNav~\cite{savva2017minos,anderson2018evaluation}, the agent is given a 2D coordinate (relative to its position) and needs to reach that target as quickly as possible. Following~\cite{chaplot2020learning}, we use noise-free evaluation and directly transfer the mapper, planner, and local policy learned during exploration to this task. In this way, instead of navigating to a point specified by the global policy, the agent has to navigate to a fixed goal location. To evaluate navigation, we use the standard metrics---success rate, success rate normalized by inverse path length (SPL)~\cite{anderson2018evaluation}, and time taken. The agent succeeds if it stops within $0.2\si{m}$ of the target under a time budget of $T=1000$.

Table~\ref{tab:pointnav_results} shows the navigation results on the Gibson validation set. Our approach outperforms the baselines. Thus, not only does occupancy anticipation successfully map the environment, but it also allows the agent to move to a specified goal more quickly by modeling the navigable spaces. This apples-to-apples comparison shows that our idea improves the state of the art for PointNav. As with exploration, using ground truth (GT) anticipation leads to good gains in the navigation performance, and our methods bridge the gap between the prior state of the art and perfect anticipation.

\begin{table}[!t]
\begin{minipage}{\linewidth}
    \centering
    \scalebox{0.77}{
    \begin{tabular}{@{}p{4cm}p{2.0cm}p{2.0cm}p{2.0cm}@{}}
    \toprule
    Method                               &       SPL \%         &       Success \%     &         Time taken        \\ \midrule
    ANS(rgb)~\cite{chaplot2020learning}  &       66.8           &       87.9           &         254.109           \\
    ANS(depth)                           &       76.8           &       86.6           &         226.161           \\ 
    View-extrap.                         &       10.4           &       33.3           &         835.556           \\ \midrule
    OccAnt(rgb)                          &       71.2           &       88.2           &         223.411           \\
    OccAnt(depth)                        &       77.8           &       91.3           &         194.751           \\
    OccAnt(rgbd)                         &   \tb{80.0}          &   \tb{93.0}          &     \tb{171.874}          \\ \midrule
    \rowcolor{Gray} OccAnt(GT)           &       89.5           &       96.0           &         125.018           \\ \bottomrule
    \end{tabular}
    }
    \caption{\small\textbf{PointNav results:} Our approach provides more efficient navigation.}
    \label{tab:pointnav_results}
\end{minipage}
\begin{minipage}{\linewidth}
    \centering
    \scalebox{0.75}{
    \setlength{\tabcolsep}{3pt}
    \begin{tabular}{@{}clp{1.5cm}p{1.5cm}lp{1.5cm}p{1.5cm}@{}}
    \toprule
         & \multicolumn{3}{c}{Test standard}                                                & \multicolumn{3}{c}{Test challenge}                                           \\ \cmidrule(lr){2-4}\cmidrule(lr){5-7}
    Rank & Team                                                       & SPL \% & Success \% & Team                                                  & SPL \%  & Success \% \\ \cmidrule(lr){2-4}\cmidrule(lr){5-7}
    1    & \tb{OccupancyAnticipation}                                 & 19.2   & 24.8       & \tb{OccupancyAnticipation}                            & 20.9    & 27.5       \\
    2    & ego-localization~\cite{datta2020egolocalization}           & 10.4   & 13.6       & ego-localization ~\cite{datta2020egolocalization}     & 14.6    & 19.2       \\
    3    & Information Bottleneck                                     & 5.0    & 7.5        & DAN~\cite{karkus2019differentiable}                   & 13.2    & 25.3       \\
    4    & cogmodel\_team                                             & 0.8    & 1.3        & Information Bottleneck                                & 6.0     & 8.8        \\
    5    & UCULab                                                     & 0.5    & 0.8        & cogmodel\_team                                        & 0.7     & 1.2        \\
    6    & Habitat Team (DD-PPO)~\cite{wijmans2019decentralized}      & 0.0    & 0.2        & UCULab                                                & 0.1     & 0.2        \\ \bottomrule
    \end{tabular}
    }
    \caption{\small \textbf{Habitat Challenge 2020 results:} Our approach is the winning entry.}
    \label{tab:habitat_challenge}
\end{minipage}
\end{table}

In concurrent work, the DD-PPO approach~\cite{wijmans2019decentralized} obtains 0.96 SPL for PointNav, but it requires 2.5 billion frames of experience to do so (and it fails for noisy conditions; see below). To achieve the performance of our method (0.8 SPL in 2M frames), DD-PPO requires more than 50$\times$ the experience. Our sample efficiency can be attributed to explicit mapping along with occupancy anticipation. 

Finally, we validate our approach on the 2020 Habitat PointNav Challenge~\cite{habitat-challenge}, which requires the agent to adapt to noisy RGB-D sensors and noisy actuators, and to operate without an odometer. This presents a much more difficult evaluation setup than past work which assumes perfect odometry as well as noise-free sensing and actuation~\cite{habitat19iccv,chaplot2020learning,wijmans2019decentralized}. See Sec.~S13 in Supp.~for more details. Table~\ref{tab:habitat_challenge} shows the results.
Our method won the challenge, outperforming the competing approaches by large margins. While our approach generalizes well to this setting, DD-PPO~\cite{wijmans2019decentralized} fails (0 SPL) due to its reliance on perfect odometry.
\section{Conclusion}

We introduced the idea of occupancy anticipation from egocentric views in 3D environments.  By learning to anticipate the navigable areas beyond the agent's actual field of view, we obtain more accurate maps more efficiently in novel environments.  We demonstrate our idea both for individual local maps, as well as integrated within sequential models for exploration and navigation, where the agent continually refines its (anticipated) map of the world.  Our results clearly demonstrate the advantages on multiple datasets, including improvements to the state-of-the-art embodied AI model for exploration and navigation.

\section*{Acknowledgements}

UT Austin is supported in part by DARPA Lifelong Learning Machines and the GCP Research Credits Program. We thank Devendra Singh Chaplot for clarifying the implementation details for ANS.

\bibliographystyle{splncs04}
\bibliography{egbib}

\begin{thebibliography}{10}
\providecommand{\url}[1]{\texttt{#1}}
\providecommand{\urlprefix}{URL }
\providecommand{\doi}[1]{https://doi.org/#1}

\bibitem{habitat-challenge}
The {H}abitat {C}hallenge 2020. \url{https://aihabitat.org/challenge/2020/ }

\bibitem{anderson2018evaluation}
Anderson, P., Chang, A., Chaplot, D.S., Dosovitskiy, A., Gupta, S., Koltun, V.,
  Kosecka, J., Malik, J., Mottaghi, R., Savva, M., et~al.: On evaluation of
  embodied navigation agents. arXiv preprint arXiv:1807.06757  (2018)

\bibitem{mattersim}
Anderson, P., Wu, Q., Teney, D., Bruce, J., Johnson, M., S{\"u}nderhauf, N.,
  Reid, I., Gould, S., van~den Hengel, A.: Vision-and-language navigation:
  Interpreting visually-grounded navigation instructions in real environments.
  In: Proceedings of the IEEE Conference on Computer Vision and Pattern
  Recognition (CVPR) (2018)

\bibitem{stanford2d3d}
{Armeni}, I., {Sax}, A., {Zamir}, A.R., {Savarese}, S.: {Joint 2D-3D-Semantic
  Data for Indoor Scene Understanding}. ArXiv e-prints  (Feb 2017)

\bibitem{bao2012semantic}
Bao, S.Y., Bagra, M., Chao, Y.W., Savarese, S.: Semantic structure from motion
  with points, regions, and objects. In: 2012 IEEE Conference on Computer
  Vision and Pattern Recognition. pp. 2703--2710. IEEE (2012)

\bibitem{burada2018curiosity}
Burda, Y., Edwards, H., Pathak, D., Storkey, A., Darrell, T., Efros, A.A.:
  Large-scale study of curiosity-driven learning. In: arXiv:1808.04355 (2018)

\bibitem{cadena2016past}
Cadena, C., Carlone, L., Carrillo, H., Latif, Y., Scaramuzza, D., Neira, J.,
  Reid, I., Leonard, J.J.: Past, present, and future of simultaneous
  localization and mapping: Toward the robust-perception age. IEEE Transactions
  on robotics  \textbf{32}(6),  1309--1332 (2016)

\bibitem{carrillo2012comparison}
Carrillo, H., Reid, I., Castellanos, J.A.: On the comparison of uncertainty
  criteria for active slam. In: 2012 IEEE International Conference on Robotics
  and Automation. pp. 2080--2087. IEEE (2012)

\bibitem{chang2017matterport}
Chang, A., Dai, A., Funkhouser, T., Nie{\ss}ner, M., Savva, M., Song, S., Zeng,
  A., Zhang, Y.: Matterport3d: Learning from rgb-d data in indoor environments.
  In: Proceedings of the International Conference on 3D Vision (3DV) (2017),
  matterPort3D dataset license available at:
  \url{http://kaldir.vc.in.tum.de/matterport/MP_TOS.pdf}.

\bibitem{chaplot2020learning}
Chaplot, D.S., Gupta, S., Gandhi, D., Gupta, A., Salakhutdinov, R.: Learning to
  explore using active neural mapping. 8th International Conference on Learning
  Representations, ICLR 2020  (2020)

\bibitem{chen2019learning}
Chen, T., Gupta, S., Gupta, A.: Learning exploration policies for navigation.
  In: 7th International Conference on Learning Representations, ICLR 2019
  (2019)

\bibitem{Choi_2015_CVPR}
Choi, S., Zhou, Q.Y., Koltun, V.: Robust reconstruction of indoor scenes. In:
  IEEE Conference on Computer Vision and Pattern Recognition (CVPR) (2015)

\bibitem{das2018embodied}
Das, A., Datta, S., Gkioxari, G., Lee, S., Parikh, D., Batra, D.: Embodied
  question answering. In: Proceedings of the IEEE Conference on Computer Vision
  and Pattern Recognition Workshops. pp. 2054--2063 (2018)

\bibitem{datta2020egolocalization}
Datta, S., Maksymets, O., Hoffman, J., Lee, S., Batra, D., Parikh, D.:
  Integrating egocentric localization for more realistic pointgoal navigation
  agents. CVPR 2020 Embodied AI Workshop  (2020)

\bibitem{Dhamo_2019_ICCV}
Dhamo, H., Navab, N., Tombari, F.: Object-driven multi-layer scene
  decomposition from a single image. In: The IEEE International Conference on
  Computer Vision (ICCV) (October 2019)

\bibitem{elhafsi2019map}
Elhafsi, A., Ivanovic, B., Janson, L., Pavone, M.: Map-predictive motion
  planning in unknown environments. arXiv preprint arXiv:1910.08184  (2019)

\bibitem{fang2019scene}
Fang, K., Toshev, A., Fei-Fei, L., Savarese, S.: Scene memory transformer for
  embodied agents in long-horizon tasks. In: Proceedings of the IEEE Conference
  on Computer Vision and Pattern Recognition. pp. 538--547 (2019)

\bibitem{gan2019look}
Gan, C., Zhang, Y., Wu, J., Gong, B., Tenenbaum, J.B.: Look, listen, and act:
  Towards audio-visual embodied navigation. arXiv preprint arXiv:1912.11684
  (2019)

\bibitem{gordon2018iqa}
Gordon, D., Kembhavi, A., Rastegari, M., Redmon, J., Fox, D., Farhadi, A.: Iqa:
  Visual question answering in interactive environments. In: Proceedings of the
  IEEE Conference on Computer Vision and Pattern Recognition. pp. 4089--4098
  (2018)

\bibitem{gupta2017cognitive}
Gupta, S., Davidson, J., Levine, S., Sukthankar, R., Malik, J.: Cognitive
  mapping and planning for visual navigation. In: Proceedings of the IEEE
  Conference on Computer Vision and Pattern Recognition. pp. 2616--2625 (2017)

\bibitem{gupta2017unifying}
Gupta, S., Fouhey, D., Levine, S., Malik, J.: Unifying map and landmark based
  representations for visual navigation. arXiv preprint arXiv:1712.08125
  (2017)

\bibitem{Hart1968}
Hart, P., Nilsson, N., Raphael, B.: A formal basis for the heuristic
  determination of minimum cost paths. {IEEE} Transactions on Systems Science
  and Cybernetics  \textbf{4}(2),  100--107 (1968).
  \doi{10.1109/tssc.1968.300136},
  \url{https://doi.org/10.1109/tssc.1968.300136}

\bibitem{hartley2003multiple}
Hartley, R., Zisserman, A.: Multiple view geometry in computer vision.
  Cambridge university press (2003)

\bibitem{henriques2018mapnet}
Henriques, J.F., Vedaldi, A.: Mapnet: An allocentric spatial memory for mapping
  environments. In: proceedings of the IEEE Conference on Computer Vision and
  Pattern Recognition. pp. 8476--8484 (2018)

\bibitem{hoermann2018dynamic}
Hoermann, S., Bach, M., Dietmayer, K.: Dynamic occupancy grid prediction for
  urban autonomous driving: A deep learning approach with fully automatic
  labeling. In: 2018 IEEE International Conference on Robotics and Automation
  (ICRA). pp. 2056--2063. IEEE (2018)

\bibitem{iizuka2017globally}
Iizuka, S., Simo-Serra, E., Ishikawa, H.: Globally and locally consistent image
  completion. ACM Transactions on Graphics (ToG)  \textbf{36}(4),  1--14 (2017)

\bibitem{jayaraman2018shapecodes}
Jayaraman, D., Gao, R., Grauman, K.: Shapecodes: self-supervised feature
  learning by lifting views to viewgrids. In: Proceedings of the European
  Conference on Computer Vision (ECCV). pp. 120--136 (2018)

\bibitem{dinesh2018ltla}
Jayaraman, D., Grauman, K.: Learning to look around: Intelligently exploring
  unseen environments for unknown tasks. In: Computer Vision and Pattern
  Recognition, 2018 IEEE Conference on (2018)

\bibitem{cjc2019floorsp}
Jiacheng~Chen, Chen~Liu, J.W., Furukawa, Y.: Floor-sp: Inverse cad for
  floorplans by sequential room-wise shortest path. In: The IEEE International
  Conference on Computer Vision (ICCV) (2019)

\bibitem{karkus2019differentiable}
Karkus, P., Ma, X., Hsu, D., Kaelbling, L.P., Lee, W.S., Lozano-P{\'e}rez, T.:
  Differentiable algorithm networks for composable robot learning. arXiv
  preprint arXiv:1905.11602  (2019)

\bibitem{katyal2019uncertainty}
Katyal, K., Popek, K., Paxton, C., Burlina, P., Hager, G.D.: Uncertainty-aware
  occupancy map prediction using generative networks for robot navigation. In:
  2019 International Conference on Robotics and Automation (ICRA). pp.
  5453--5459. IEEE (2019)

\bibitem{katyal2018occupancy}
Katyal, K., Popek, K., Paxton, C., Moore, J., Wolfe, K., Burlina, P., Hager,
  G.D.: Occupancy map prediction using generative and fully convolutional
  networks for vehicle navigation. arXiv preprint arXiv:1803.02007  (2018)

\bibitem{kingma2014adam}
Kingma, D.P., Ba, J.: Adam: A method for stochastic optimization. arXiv
  preprint arXiv:1412.6980  (2014)

\bibitem{ai2thor}
Kolve, E., Mottaghi, R., Han, W., VanderBilt, E., Weihs, L., Herrasti, A.,
  Gordon, D., Zhu, Y., Gupta, A., Farhadi, A.: {AI2-THOR: An Interactive 3D
  Environment for Visual AI}. arXiv  (2017)

\bibitem{li2017generative}
Li, Y., Liu, S., Yang, J., Yang, M.H.: Generative face completion. In:
  Proceedings of the IEEE Conference on Computer Vision and Pattern
  Recognition. pp. 3911--3919 (2017)

\bibitem{liu2018floornet}
Liu, C., Wu, J., Furukawa, Y.: Floornet: A unified framework for floorplan
  reconstruction from 3d scans. In: Proceedings of the European Conference on
  Computer Vision (ECCV). pp. 201--217 (2018)

\bibitem{lu2019hallucinating}
Lu, C., Dubbelman, G.: Hallucinating beyond observation: Learning to complete
  with partial observation and unpaired prior knowledge (2019)

\bibitem{habitat19iccv}
{Manolis Savva*}, {Abhishek Kadian*}, {Oleksandr Maksymets*}, Zhao, Y.,
  Wijmans, E., Jain, B., Straub, J., Liu, J., Koltun, V., Malik, J., Parikh,
  D., Batra, D.: Habitat: {A} {P}latform for {E}mbodied {AI} {R}esearch. In:
  Proceedings of the IEEE/CVF International Conference on Computer Vision
  (ICCV) (2019)

\bibitem{martinez2009bayesian}
Martinez-Cantin, R., De~Freitas, N., Brochu, E., Castellanos, J., Doucet, A.: A
  bayesian exploration-exploitation approach for optimal online sensing and
  planning with a visually guided mobile robot. Autonomous Robots
  \textbf{27}(2),  93--103 (2009)

\bibitem{mohajerin2019multi}
Mohajerin, N., Rohani, M.: Multi-step prediction of occupancy grid maps with
  recurrent neural networks. In: Proceedings of the IEEE Conference on Computer
  Vision and Pattern Recognition. pp. 10600--10608 (2019)

\bibitem{mousavian2019visual}
Mousavian, A., Toshev, A., Fi{\v{s}}er, M., Ko{\v{s}}eck{\'a}, J., Wahid, A.,
  Davidson, J.: Visual representations for semantic target driven navigation.
  In: 2019 International Conference on Robotics and Automation (ICRA). pp.
  8846--8852. IEEE (2019)

\bibitem{muller2018driving}
M{\"u}ller, M., Dosovitskiy, A., Ghanem, B., Koltun, V.: Driving policy
  transfer via modularity and abstraction. arXiv preprint arXiv:1804.09364
  (2018)

\bibitem{pyrobot2019}
Murali, A., Chen, T., Alwala, K.V., Gandhi, D., Pinto, L., Gupta, S., Gupta,
  A.: Pyrobot: An open-source robotics framework for research and benchmarking.
  arXiv preprint arXiv:1906.08236  (2019)

\bibitem{odena2016deconvolution}
Odena, A., Dumoulin, V., Olah, C.: Deconvolution and checkerboard artifacts.
  Distill  \textbf{1}(10), ~e3 (2016)

\bibitem{o2012gaussian}
O’Callaghan, S.T., Ramos, F.T.: Gaussian process occupancy maps. The
  International Journal of Robotics Research  \textbf{31}(1),  42--62 (2012)

\bibitem{parisotto2017neural}
Parisotto, E., Salakhutdinov, R.: Neural map: Structured memory for deep
  reinforcement learning. arXiv preprint arXiv:1702.08360  (2017)

\bibitem{NEURIPS2019_9015}
Paszke, A., Gross, S., Massa, F., Lerer, A., Bradbury, J., Chanan, G., Killeen,
  T., Lin, Z., Gimelshein, N., Antiga, L., Desmaison, A., Kopf, A., Yang, E.,
  DeVito, Z., Raison, M., Tejani, A., Chilamkurthy, S., Steiner, B., Fang, L.,
  Bai, J., Chintala, S.: Pytorch: An imperative style, high-performance deep
  learning library. In: Advances in Neural Information Processing Systems 32,
  pp. 8024--8035. Curran Associates, Inc. (2019),
  \url{http://papers.neurips.cc/paper/9015-pytorch-an-imperative-style-high-performance-deep-learning-library.pdf}

\bibitem{pathak2017curiosity}
Pathak, D., Agrawal, P., Efros, A.A., Darrell, T.: Curiosity-driven exploration
  by self-supervised prediction. In: International Conference on Machine
  Learning (2017)

\bibitem{Pathak_2016_CVPR}
Pathak, D., Krahenbuhl, P., Donahue, J., Darrell, T., Efros, A.A.: Context
  encoders: Feature learning by inpainting. In: The IEEE Conference on Computer
  Vision and Pattern Recognition (CVPR) (June 2016)

\bibitem{ramakrishnan2018sidekick}
Ramakrishnan, S.K., Grauman, K.: Sidekick policy learning for active visual
  exploration. In: Proceedings of the European Conference on Computer Vision
  (ECCV). pp. 413--430 (2018)

\bibitem{ramakrishnan2019emergence}
Ramakrishnan, S.K., Jayaraman, D., Grauman, K.: Emergence of exploratory
  look-around behaviors through active observation completion. Science Robotics
   \textbf{4}(30) (2019). \doi{10.1126/scirobotics.aaw6326},
  \url{https://robotics.sciencemag.org/content/4/30/eaaw6326}

\bibitem{ramakrishnan2020exploration}
Ramakrishnan, S.K., Jayaraman, D., Grauman, K.: An exploration of embodied
  visual exploration. arXiv preprint arXiv:2001.02192  (2020)

\bibitem{ramos2016hilbert}
Ramos, F., Ott, L.: Hilbert maps: scalable continuous occupancy mapping with
  stochastic gradient descent. The International Journal of Robotics Research
  \textbf{35}(14),  1717--1730 (2016)

\bibitem{ronneberger2015u}
Ronneberger, O., Fischer, P., Brox, T.: U-net: Convolutional networks for
  biomedical image segmentation. In: International Conference on Medical image
  computing and computer-assisted intervention. pp. 234--241. Springer (2015)

\bibitem{salas2013slam++}
Salas-Moreno, R.F., Newcombe, R.A., Strasdat, H., Kelly, P.H., Davison, A.J.:
  Slam++: Simultaneous localisation and mapping at the level of objects. In:
  Proceedings of the IEEE conference on computer vision and pattern
  recognition. pp. 1352--1359 (2013)

\bibitem{savinov2018semi}
Savinov, N., Dosovitskiy, A., Koltun, V.: Semi-parametric topological memory
  for navigation. arXiv preprint arXiv:1803.00653  (2018)

\bibitem{savinov2018episodic}
Savinov, N., Raichuk, A., Marinier, R., Vincent, D., Pollefeys, M., Lillicrap,
  T., Gelly, S.: Episodic curiosity through reachability. arXiv preprint
  arXiv:1810.02274  (2018)

\bibitem{savva2017minos}
Savva, M., Chang, A.X., Dosovitskiy, A., Funkhouser, T., Koltun, V.: Minos:
  Multimodal indoor simulator for navigation in complex environments. arXiv
  preprint arXiv:1712.03931  (2017)

\bibitem{sax2018mid}
Sax, A., Emi, B., Zamir, A.R., Guibas, L., Savarese, S., Malik, J.: Mid-level
  visual representations improve generalization and sample efficiency for
  learning visuomotor policies. arXiv preprint arXiv:1812.11971  (2018)

\bibitem{schulman2017proximal}
Schulman, J., Wolski, F., Dhariwal, P., Radford, A., Klimov, O.: Proximal
  policy optimization algorithms. arXiv preprint arXiv:1707.06347  (2017)

\bibitem{seifi2019look}
Seifi, S., Tuytelaars, T.: Where to look next: Unsupervised active visual
  exploration on 360 $\{\backslash$deg$\}$ input. arXiv preprint
  arXiv:1909.10304  (2019)

\bibitem{senanayake2017deep}
Senanayake, R., Ganegedara, T., Ramos, F.: Deep occupancy maps: a continuous
  mapping technique for dynamic environments  (2017)

\bibitem{sethian1996fast}
Sethian, J.A.: A fast marching level set method for monotonically advancing
  fronts. Proceedings of the National Academy of Sciences  \textbf{93}(4),
  1591--1595 (1996)

\bibitem{shen2019situational}
Shen, W.B., Xu, D., Zhu, Y., Guibas, L.J., Fei-Fei, L., Savarese, S.:
  Situational fusion of visual representation for visual navigation. In:
  Proceedings of the IEEE International Conference on Computer Vision. pp.
  2881--2890 (2019)

\bibitem{shrestha2019learned}
Shrestha, R., Tian, F.P., Feng, W., Tan, P., Vaughan, R.: Learned map
  prediction for enhanced mobile robot exploration. In: 2019 International
  Conference on Robotics and Automation (ICRA). pp. 1197--1204. IEEE (2019)

\bibitem{sless2019self}
Sless, L., Cohen, G., Shlomo, B.E., Oron, S.: Self supervised occupancy grid
  learning from sparse radar for autonomous driving. arXiv preprint
  arXiv:1904.00415  (2019)

\bibitem{song2016ssc}
Song, S., Yu, F., Zeng, A., Chang, A.X., Savva, M., Funkhouser, T.: Semantic
  scene completion from a single depth image. Proceedings of 30th IEEE
  Conference on Computer Vision and Pattern Recognition  (2017)

\bibitem{song2018im2pano3d}
Song, S., Zeng, A., Chang, A.X., Savva, M., Savarese, S., Funkhouser, T.:
  Im2pano3d: Extrapolating 360 structure and semantics beyond the field of
  view. In: Proceedings of the IEEE Conference on Computer Vision and Pattern
  Recognition. pp. 3847--3856 (2018)

\bibitem{replica19arxiv}
Straub, J., Whelan, T., Ma, L., Chen, Y., Wijmans, E., Green, S., Engel, J.J.,
  Mur-Artal, R., Ren, C., Verma, S., Clarkson, A., Yan, M., Budge, B., Yan, Y.,
  Pan, X., Yon, J., Zou, Y., Leon, K., Carter, N., Briales, J., Gillingham, T.,
  Mueggler, E., Pesqueira, L., Savva, M., Batra, D., Strasdat, H.M., Nardi,
  R.D., Goesele, M., Lovegrove, S., Newcombe, R.: The {R}eplica dataset: A
  digital replica of indoor spaces. arXiv preprint arXiv:1906.05797  (2019)

\bibitem{Sun_2019_CVPR}
Sun, C., Hsiao, C.W., Sun, M., Chen, H.T.: Horizonnet: Learning room layout
  with 1d representation and pano stretch data augmentation. In: The IEEE
  Conference on Computer Vision and Pattern Recognition (CVPR) (June 2019)

\bibitem{thrun2002probabilistic}
Thrun, S.: Probabilistic robotics. Communications of the ACM  \textbf{45}(3),
  52--57 (2002)

\bibitem{wijmans2019decentralized}
Wijmans, E., Kadian, A., Morcos, A., Lee, S., Essa, I., Parikh, D., Savva, M.,
  Batra, D.: Dd-ppo: Learning near-perfect pointgoal navigators from 2.5
  billion frames (2020)

\bibitem{wu2019residential}
Wu, W., Fu, X.M., Tang, R., Wang, Y., Qi, Y.H., Liu, L.: Data-driven interior
  plan generation for residential buildings. ACM Trans. Graph.  \textbf{38}(6)
  (Nov 2019). \doi{10.1145/3355089.3356556},
  \url{https://doi.org/10.1145/3355089.3356556}

\bibitem{xia2018gibson}
Xia, F., Zamir, A.R., He, Z., Sax, A., Malik, J., Savarese, S.: Gibson env:
  Real-world perception for embodied agents. In: Proceedings of the IEEE
  Conference on Computer Vision and Pattern Recognition. pp. 9068--9079 (2018),
  gibson dataset license agreement available at
  \url{https://storage.googleapis.com/gibson_material/Agreement%20GDS%2006-04-18.pdf}

\bibitem{amodal}
Yang, J., Ren, Z., Xu, M., Chen, X., Crandall, D., Parikh, D., Batra, D.:
  Embodied amodal recognition: Learning to move to perceive objects. In: ICCV
  (2019)

\bibitem{yang2019dula}
Yang, S.T., Wang, F.E., Peng, C.H., Wonka, P., Sun, M., Chu, H.K.: Dula-net: A
  dual-projection network for estimating room layouts from a single rgb
  panorama. In: Proceedings of the IEEE Conference on Computer Vision and
  Pattern Recognition. pp. 3363--3372 (2019)

\bibitem{yang2018visual}
Yang, W., Wang, X., Farhadi, A., Gupta, A., Mottaghi, R.: Visual semantic
  navigation using scene priors. arXiv preprint arXiv:1810.06543  (2018)

\bibitem{Yang_2019_CVPR}
Yang, Z., Pan, J.Z., Luo, L., Zhou, X., Grauman, K., Huang, Q.: Extreme
  relative pose estimation for rgb-d scans via scene completion. In: The IEEE
  Conference on Computer Vision and Pattern Recognition (CVPR) (June 2019)

\bibitem{zhu-iccv2017}
Zhu, Y., Gordon, D., Kolve, E., Fox, D., Fei-Fei, L., Gupta, A., Mottaghi, R.,
  Farhadi, A.: {Visual Semantic Planning using Deep Successor Representations}.
  In: Computer Vision, 2017 IEEE International Conference on (2017)

\bibitem{zou2018layoutnet}
Zou, C., Colburn, A., Shan, Q., Hoiem, D.: Layoutnet: Reconstructing the 3d
  room layout from a single rgb image. In: Proceedings of the IEEE Conference
  on Computer Vision and Pattern Recognition. pp. 2051--2059 (2018)

\end{thebibliography}

\vfill
\pagebreak

\title{Occupancy Anticipation for Efficient Exploration and Navigation \\ Supplementary Materials}
\titlerunning{Supplementary Materials}

\author{Santhosh K. Ramakrishnan\inst{1,2} \and
Ziad Al-Halah\inst{1} \and
Kristen Grauman\inst{1,2}}

\authorrunning{S. Ramakrishnan et al.}

\institute{The University of Texas at Austin, Austin TX 78712, USA \and
Facebook AI Research, Austin TX 78701, USA \\
\email{srama@cs.utexas.edu,~ziadlhlh@gmail.com,~grauman@cs.utexas.edu}\\
}

\newpage
\clearpage
\maketitle
\setcounter{section}{0}
\setcounter{figure}{0}
\setcounter{table}{0}
\renewcommand{\thesection}{S\arabic{section}}
\renewcommand{\thetable}{S\arabic{table}}
\renewcommand{\thefigure}{S\arabic{figure}}

This document provides additional information about the experimental settings as well as qualitative and quantitative results to support the experiments from the main paper. Below is a summary of the sections in the supplementary file:
\begin{itemize}
    \item (\S\ref{sec:supp_noise_free}) Noise-free exploration results
    \item (\S\ref{sec:occupancy_anticipation_ablation}) Occupancy anticipation ablation study
    \item (\S\ref{sec:occupancy_anticipation_qual_examples}) Occupancy anticipation qualitative examples
    \item (\S\ref{sec:supp_exploration_qualitative}) Exploration with occupancy anticipation examples
    \item (\S\ref{sec:ground_truth_generation}) Generating ground-truth for occupancy anticipation
    \item (\S\ref{sec:noise_models}) Noise models for actuation and odometry
    \item (\S\ref{sec:differences_ans}) Differences in ANS implementation
    \item (\S\ref{sec:implementation_details}) Implementation details
    \item (\S\ref{sec:occupancy_anticipation_architecture}) Occupancy anticipation architecture
    \item (\S\ref{sec:view_extrapolation_baseline}) View extrapolation baseline
    \item (\S\ref{sec:model_capacity_comparison}) Comparing the model capacities of different methods
    \item (\S\ref{sec:habitat_challenge_2020}) Habitat challenge 2020
\end{itemize}

\section{Noise-free exploration results}
\label{sec:supp_noise_free}
\begin{figure}[ht]
    \centering
    \includegraphics[width=1.0\textwidth,trim=0 1.0cm 6.1cm 0]{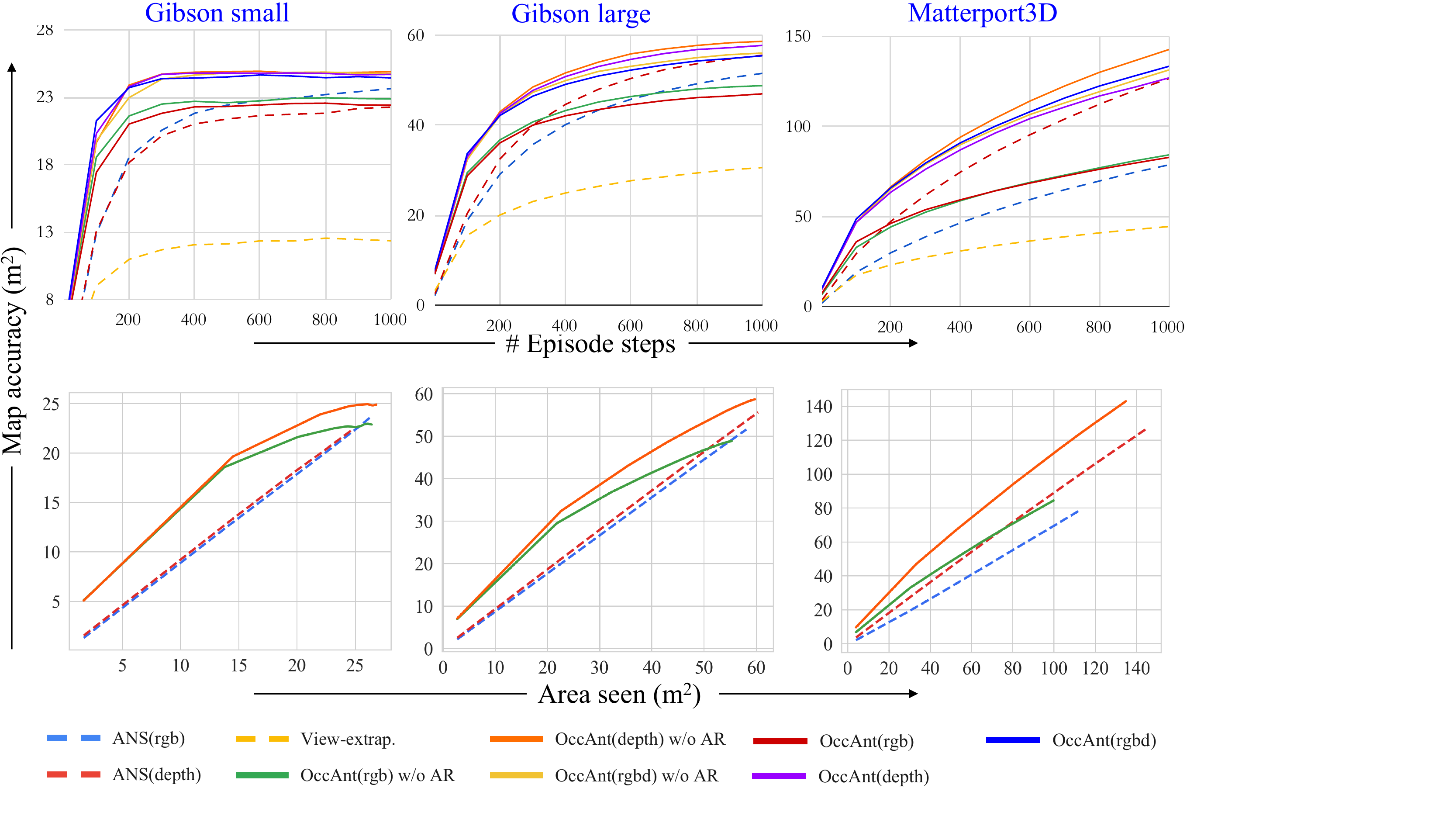}
    \caption{\small\textbf{Noise-free exploration results:} Map accuracy (\si{m^2}) as a function of episode duration (top row) and area seen (bottom row) for Gibson (small and large splits) and Matterport3D under noise-free conditions. \textbf{Top:} Our OccAnt approach (solid lines) rapidly attains higher map accuracy than the baselines (dotted lines). Using anticipation reward (AR) largely retains the original performance in the noise-free conditions (but improves significantly in the noisy conditions, see Fig.~5 main paper). \textbf{Bottom:} OccAnt achieves higher map accuracy for the same area covered (we show best variants here to avoid clutter). These results show the agent \emph{actively moves better} to explore the environment with our occupancy anticipation idea.}
    \label{fig:noise_free_exploration_results}
\end{figure}

As noted in the main paper, we evaluate on both noisy and noise-free conditions.  We showed the change in map accuracy as a function of episode steps and area seen under noisy conditions in Fig.~\ref{fig:exploration_results} in the main paper. We show the same results on noise-free conditions here in Fig.~\ref{fig:noise_free_exploration_results}. Similar to the noisy case, OccAnt approach (solid lines) rapidly leads to higher map accuracy when compared to the baselines (dotted lines). However, we can see that adding the anticipation reward (AR) in this noise-free setting does not lead to improvements in performance in contrast to what was observed for the more realistic noisy setup (Fig. 5 in main). 

As we will qualitatively demonstrate in Sec.~\ref{sec:supp_exploration_qualitative}, the main benefit of using the anticipation reward is that it leads to better noise correction in the pose estimates under noisy test conditions, resulting in more effective map registration. This is due to the fact that achieving high AR (i.e., the map accuracy) inherently depends on better map registration. If the per-frame maps are not registered correctly, AR is likely to be low even if the per-frame map estimates are very good. Therefore, in addition to covering more area, the agent also has to better train the pose estimator which would then lead to higher AR over time. Since noise correction is not needed under noise-free conditions, using AR has limited impact on the final performance.

\section{Occupancy anticipation ablation study}
\label{sec:occupancy_anticipation_ablation}

As discussed in the main paper, our key contributions are a novel framework for occupancy anticipation and a novel anticipation reward which encourages the agent to build more accurate maps (as opposed to covering more area). To isolate the gains achieved by these individual contributions, we view the results from the main paper (Tables~\ref{tab:occ_anticipation},~\ref{tab:exploration_results}, and~\ref{tab:pointnav_results} in main paper) in a different way. We first group the results based on the modality (rgb/depth/rgbd), and further sort the methods based on whether they use occupancy anticipation (OccAnt) or the anticipation reward (AR).  We present these ablations for the per-frame map evaluation (Table~\ref{tab:supp_occ_anticipation}), the exploration evaluation (Table~\ref{tab:supp_exploration_results}), and the navigation evaluation (Table~\ref{tab:supp_pointnav_results}). By default, the ANS baselines do not use occupancy anticipation or the anticipation reward and our methods always use occupancy anticipation.

For per-frame maps, in Table~\ref{tab:supp_occ_anticipation} we see that adding occupancy anticipation to the base model significantly improves the IoU and F1 scores as expected. Adding the anticipation reward leads to comparable or better results, showing that it leads to better training of the mapper during the exploration training.

For exploration, in Table~\ref{tab:supp_exploration_results} we see that adding occupancy anticipation generally leads to better map quality than ANS across different modalities and testing conditions. Adding the anticipation reward (AR) leads to significant improvements in the map quality under noisy conditions for both depth and rgbd modalities (rgb slightly underperforms). This is primarily due to improved training of the mapper module which leads to better map registration (see Sec.~\ref{sec:supp_exploration_qualitative}). As we also noted in Sec.~\ref{sec:supp_noise_free}, using AR in noise-free conditions has limited impact on the performance as the pose-estimation is assumed to be perfect in these cases. It mainly benefits exploration in the more real-world testing scenarios with noisy actuation and sensing.

For navigation, in Table~\ref{tab:supp_pointnav_results} we see that adding occupancy anticipation leads to significant improvements in all three metrics. The impact of using AR here is limited because we assume noise-free test conditions for PointNav (following~\cite{habitat19iccv,chaplot2020learning}). However, the challenge results reported in the main paper remove this assumption to test PointNav with noisy odometry and actuation.

\begin{table}[ht!]
\begin{minipage}{\linewidth}
\centering
\scalebox{0.78}{
\setlength{\tabcolsep}{3pt}
\begin{tabular}{@{}p{3.5cm}P{1.3cm}P{0.7cm}P{1.1cm}P{1.1cm}P{1.1cm}P{1.1cm}P{1.1cm}P{1.1cm}@{}}
\toprule
\multirow{2}{*}{Method} &               &          &      \multicolumn{3}{c}{IoU \%}    &   \multicolumn{3}{c}{F1 score \%}    \\ \cmidrule(r){4-6} \cmidrule(r){7-9}
                        &      OccAnt   &   AR     &     free   &     occ.  &     mean  &     free   &     occ.   &     mean   \\ \midrule
 ANS(rgb)               &      \xmark   &  \xmark  &     12.1   &     14.9  &     13.5  &     19.6    &     24.9  &     22.5   \\
 OccAnt(rgb) w/o AR     &      \cmark   &  \xmark  & \tb{44.6}  &\tb{47.9}  & \tb{46.2} & \tb{58.4}   & \tb{62.9} & \tb{60.6}  \\
 OccAnt(rgb)            &      \cmark   &  \cmark  &     44.4   &\tb{47.9}  &     46.1  &     58.2    & \tb{62.9} & \tb{60.6}  \\ \midrule
 ANS(depth)             &      \xmark   &  \xmark  &     14.5   &     24.1  &     19.3  &     23.1    &     37.6  &     30.4   \\
 OccAnt(depth) w/o AR   &      \cmark   &  \xmark  &     50.3   &     61.7  &     56.0  & \tb{63.8}   &     74.9  &     69.3   \\
 OccAnt(depth)          &      \cmark   &  \cmark  & \tb{50.4}  & \tb{61.9} & \tb{56.1} & \tb{63.8}   & \tb{75.0} & \tb{69.4}  \\ \midrule
 OccAnt(rgbd) w/o AR    &      \cmark   &  \xmark  &     50.1   &     60.5  &     55.3  &     63.6    &     74.1  &     68.8   \\
 OccAnt(rgbd)           &      \cmark   &  \cmark  & \tb{51.5}  & \tb{61.5} & \tb{56.5} & \tb{64.9}   & \tb{74.8} & \tb{69.8}  \\
 \bottomrule
\end{tabular}
}

\caption{\small Per-frame occupancy anticipation ablation study}
\label{tab:supp_occ_anticipation}
\end{minipage}

\begin{minipage}{\linewidth}
\centering
\scalebox{0.78}{
\setlength{\tabcolsep}{3pt}
\begin{tabular}{@{}lP{1.3cm}P{0.7cm}P{1.3cm}P{1.3cm}P{1.3cm}P{1.3cm}P{1.3cm}P{1.3cm}@{}}
\toprule
\rowcolor{Gray}                                                   \multicolumn{9}{c}{Noisy test conditions}                                                    \\ \midrule
                               &             &        & \multicolumn{2}{c}{Gibson small} & \multicolumn{2}{c}{Gibson large} & \multicolumn{2}{c}{Matterport3D} \\ \cmidrule(r){4-5} \cmidrule(r){6-7} \cmidrule(r){8-9}
Method                         & OccAnt      & AR     &      Map acc. &       IoU \%     &      Map acc.   &     IoU \%     &      Map acc.    &      IoU \%   \\ \midrule
ANS(rgb)~\cite{chaplot2020learning} & \xmark & \xmark &     18.46     &        55        &     34.95       &       47       &     44.70        &     18        \\
OccAnt(rgb) w/o AR             &      \cmark & \xmark & \tb{21.77}    &    \tb{66}       & \tb{44.15}      &   \tb{57}      &     65.76        & \tb{23}       \\
OccAnt(rgb)                    &      \cmark & \cmark &     20.87     &        62        &     42.08       &       54       & \tb{66.15}       &     22        \\ \midrule
ANS(depth)                     &      \xmark & \xmark &     18.54     &        56        &     39.35       &       53       &     72.48        &     26        \\
OccAnt(depth) w/o AR           &      \cmark & \xmark &     20.22     &        58        &     44.18       &       54       &     92.70        &     29        \\
OccAnt(depth)                  &      \cmark & \cmark & \tb{22.74}    &    \tb{71}       & \tb{50.30}      &   \tb{67}      & \tb{94.12}       & \tb{33}       \\  \midrule
OccAnt(rgbd) w/o AR            &      \cmark & \xmark &     16.92     &        45        &     35.60       &       40       &     76.32        &     23        \\
OccAnt(rgbd)                   &      \cmark & \cmark & \tb{22.70}    &    \tb{71}       & \tb{48.42}      &   \tb{62}      & \tb{99.92}       & \tb{32}       \\ \midrule
\rowcolor{Gray}                                                   \multicolumn{9}{c}{Noise-free test conditions}                                               \\ \midrule
                               &             &        & \multicolumn{2}{c}{Gibson small} & \multicolumn{2}{c}{Gibson large} & \multicolumn{2}{c}{Matterport3D} \\ \cmidrule(r){4-5} \cmidrule(r){6-7} \cmidrule(r){8-9}
Method                         & OccAnt      & AR     &    Map acc.   &       IoU \%     &      Map acc.   &     IoU \%     &      Map acc.    &      IoU \%   \\ \midrule
ANS(rgb)~\cite{chaplot2020learning} & \xmark & \xmark &     22.43     &    \tb{76}       &     43.41       &   \tb{64}      &     53.40        &     23        \\
OccAnt(rgb) w/o AR             &      \cmark & \xmark & \tb{22.60}    &        71        & \tb{45.19}      &       60       & \tb{64.44}       & \tb{24}       \\
OccAnt(rgb)                    &      \cmark & \cmark &     22.32     &        70        &     43.52       &       58       &     64.35        &     22        \\ \midrule
ANS(depth)                     &      \xmark & \xmark &     21.39     &        74        &     48.01       &       72       &     85.91        &     34        \\
OccAnt(depth) w/o AR           &      \cmark & \xmark & \tb{24.91}    &    \tb{84}       & \tb{54.05}      &   \tb{75}      &\tb{104.68}       & \tb{38}       \\
OccAnt(depth)                  &      \cmark & \cmark &     24.80     &        83        &     53.08       &       74       &     96.45        &     35        \\  \midrule
OccAnt(rgbd) w/o AR            &      \cmark & \xmark & \tb{24.80}    &    \tb{84}       & \tb{51.99}      &   \tb{71}      &     98.70        &     34        \\
OccAnt(rgbd)                   &      \cmark & \cmark &     24.51     &        82        &     50.97       &       69       &\tb{100.25}       &     34        \\ \bottomrule
\end{tabular}
}
\caption{\small\textbf{Timed exploration ablation:} Map quality at $T$=500 for all models and datasets.}
\label{tab:supp_exploration_results}
\end{minipage}

\begin{minipage}{\linewidth}
\centering
\scalebox{0.78}{
\setlength{\tabcolsep}{4pt}
\begin{tabular}{@{}p{3.5cm}P{1.3cm}P{0.7cm}p{1.3cm}p{2.3cm}p{2.0cm}@{}}
\toprule
Method                               &  OccAnt   &   AR     &     SPL \%         &   Success Rate \%    &         Time taken        \\ \midrule
ANS(rgb)~\cite{chaplot2020learning}  &  \xmark   &  \xmark  &     66.8           &       87.9           &         254.109           \\
OccAnt(rgb) w/o AR                   &  \cmark   &  \xmark  & \tb{71.2}          &   \tb{88.2}          &     \tb{223.411}          \\
OccAnt(rgb)                          &  \cmark   &  \cmark  &     66.1           &       81.3           &         293.321           \\ \midrule
ANS(depth)                           &  \xmark   &  \xmark  &     76.8           &       86.6           &         226.161           \\
OccAnt(depth) w/o AR                 &  \cmark   &  \xmark  & \tb{78.6}          &   \tb{92.2}          &     \tb{187.358}          \\
OccAnt(depth)                        &  \cmark   &  \cmark  &     77.8           &       91.3           &         194.751           \\ \midrule
OccAnt(rgbd) w/o AR                  &  \cmark   &  \xmark  &     77.9           &       92.9           &         174.105           \\
OccAnt(rgbd)                         &  \cmark   &  \cmark  & \tb{80.0}          &   \tb{93.0}          &     \tb{171.874}          \\ \bottomrule
\end{tabular}
}
\caption{\small\textbf{PointGoal navigation ablation:} Time taken refers to the average number of agent actions required; the maximum time budget is $T$=1000.}
\label{tab:supp_pointnav_results}
\vspace{-0.2in}
\end{minipage}
\end{table}

\section{Occupancy anticipation qualitative examples}
\label{sec:occupancy_anticipation_qual_examples}
\begin{figure}
    \centering
    \includegraphics[width=0.9\textwidth]{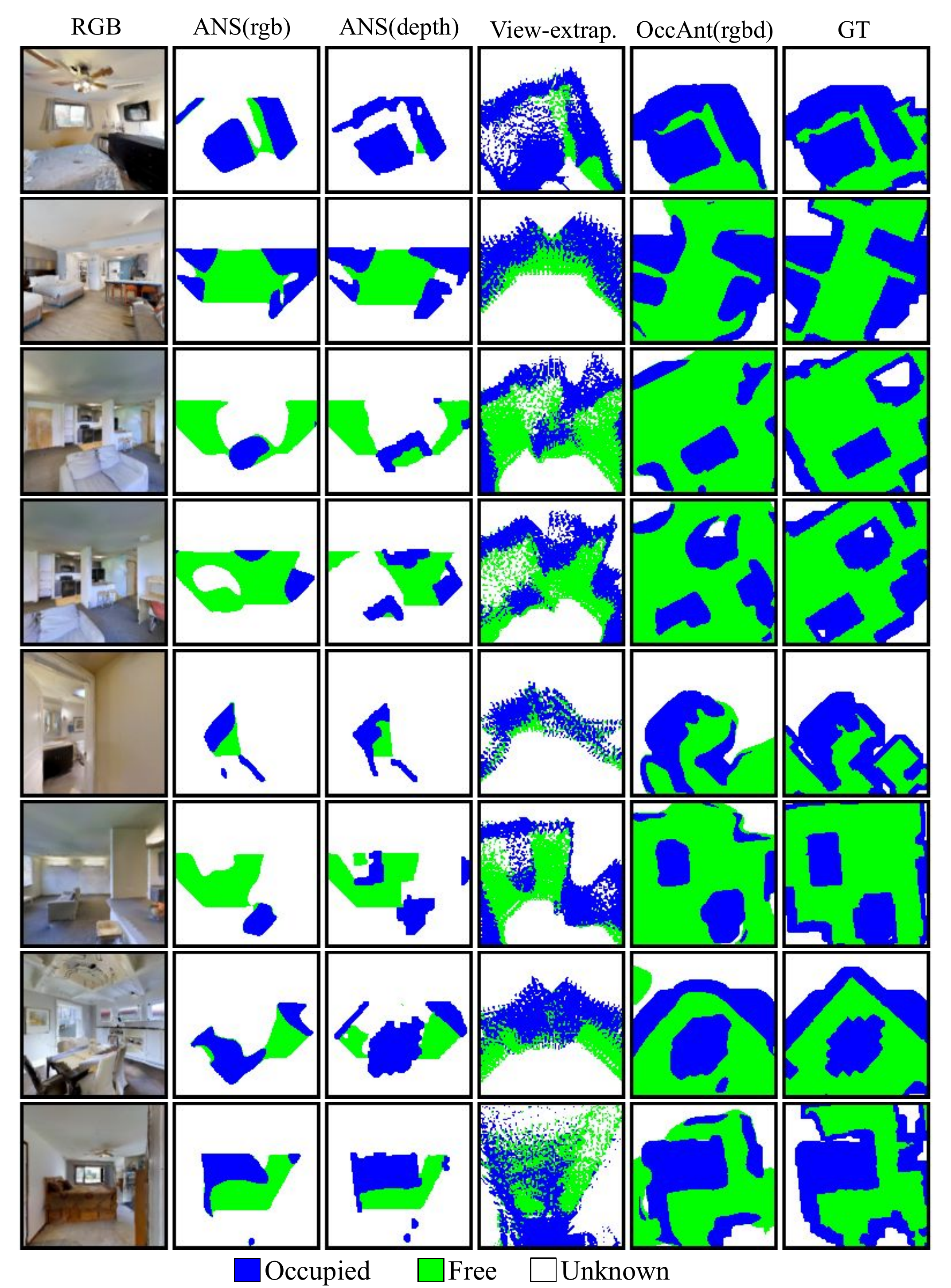}
    \caption{\small \textbf{Occupancy anticipation successful cases:} ANS(rgb) is trained to predict the visible occupancy (2nd column) and ANS(depth) (3rd column) directly uses the visible occupancy (within a $3\si{m}$ range). Both these methods are unable to account for regions that are not visible or outside the sensing range. While View-extrap (4th column) is able to expand beyond a $90^\circ$ FoV, its predictions are often noisy and do not include occluded regions. Also, the predictions are not guaranteed to be smooth in the top-down projection as smoothness in the depth-image prediction space does not necessarily lead to smoothness in the top-down maps, resulting in speckled outputs. Our method OccAnt(rgbd) (5th column) is able to successfully anticipate occupancy for regions that are occluded and outside the field-of-view with high accuracy (see ground-truth in column 6).}
    \label{fig:occ_ant_positives}
\end{figure}

\begin{figure}
    \centering
    \includegraphics[width=0.9\textwidth]{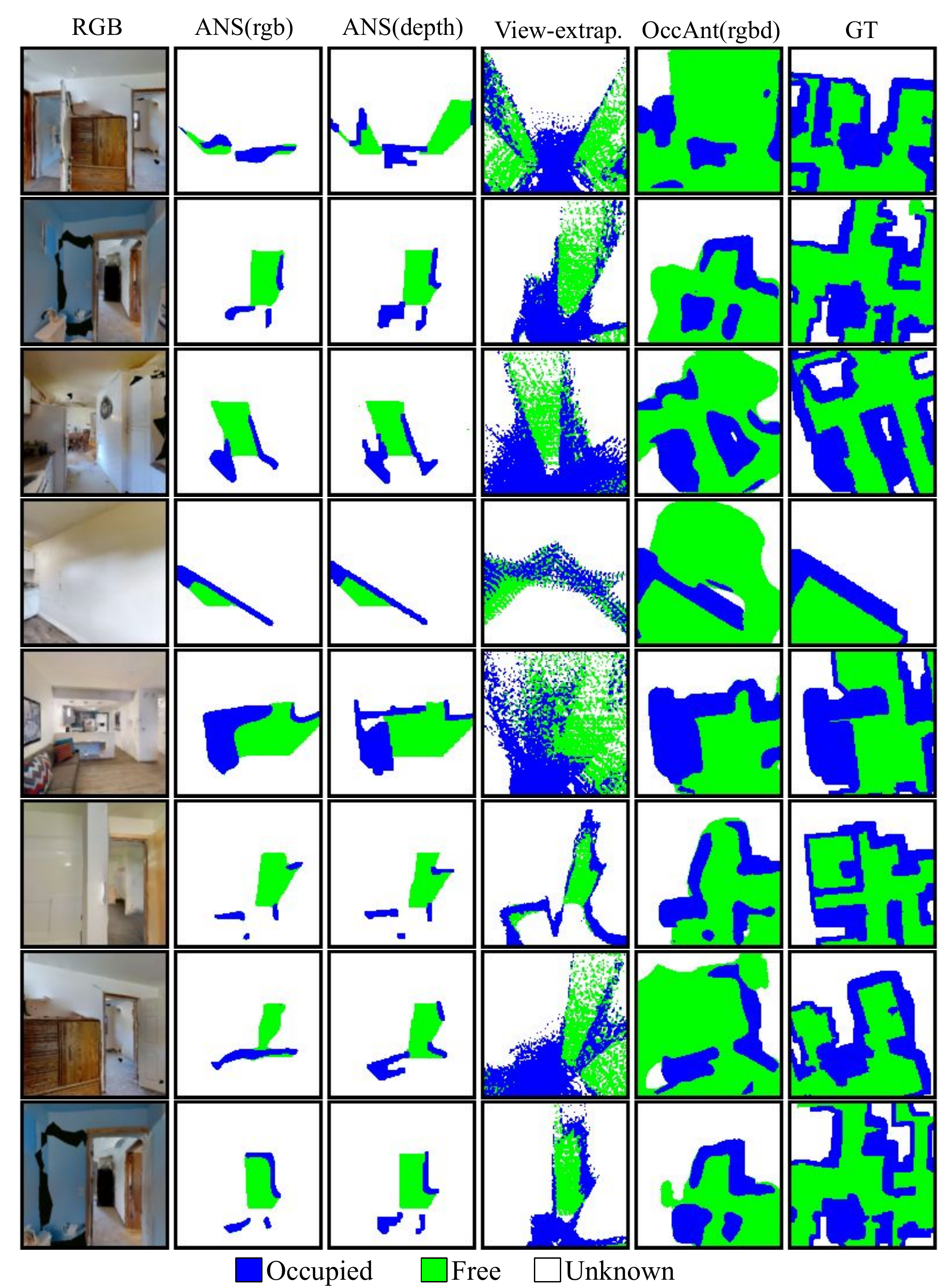}
    \caption{\small \textbf{Occupancy anticipation failure cases:} Our approach OccAnt(rgbd) incorrectly predicts narrow corridors as occupied, and is unable to handle cases where multiple solutions may exist. For example, in rows 2, 5 and 8, it predicts that the corridors in the center of the map are blocked. In row 1, it predicts that the two doors correspond to the same room, even though the wall colors are different and it is unlikely that a small room would have two doors. In row 3, 4 and 6, it predicts entrances to spaces that do not exist. Such predictions are generally difficult to make given only the context of the current first-person view, and therefore our model tends to fail at these cases.}
    \label{fig:occ_ant_negatives}
\end{figure}

See Figs.~\ref{fig:occ_ant_positives} and~\ref{fig:occ_ant_negatives} for some successful cases and failure cases for our best method from Table~\ref{tab:supp_occ_anticipation} when compared with the baselines.

\section{Exploration with occupancy anticipation examples}
\label{sec:supp_exploration_qualitative}
In Table~\ref{tab:exploration_results} and Fig.~\ref{fig:exploration_results} from the main paper, and Table~\ref{tab:supp_exploration_results} in this supplementary, we see that adding occupancy anticipation on top of the ANS baseline leads to better performance, and adding anticipation reward (AR) leads to better mapping in the noisy cases.

\begin{figure}
    \centering
    \includegraphics[width=0.75\textwidth]{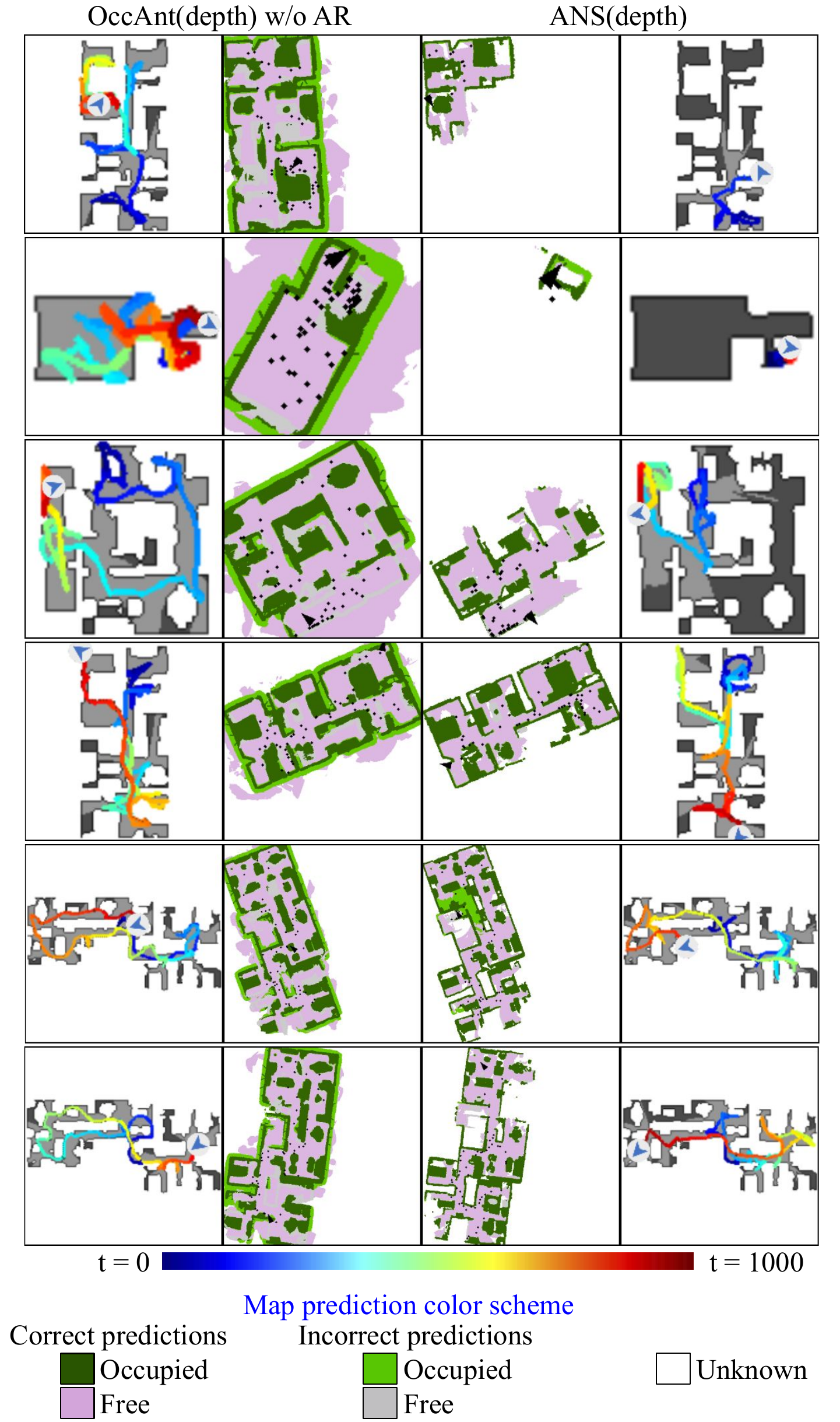}
    \caption{\small We enumerate some of the key advantages of exploration using occupancy anticipation by comparing OccAnt(depth) w/o AR with ANS(depth) in Gibson under noise-free conditions. The exploration trajectories and the map created during exploration are shown at the extremes and the center, respectively. `ANS(depth) tends to achieve worse exploration in some cases where the visible occupancy is incorrectly estimated (top 3 rows), causing the agent to get stuck in local regions. In other cases, the map accuracy is generally higher for OccAnt(depth) w/o AR for similar amounts of area seen (bottom 3 rows) as it is better at filling up the occupancy for unvisited regions. }
    \label{fig:exp_occant_depth_strengths}
\end{figure}

\begin{figure}
    \centering
    \includegraphics[width=0.75\textwidth]{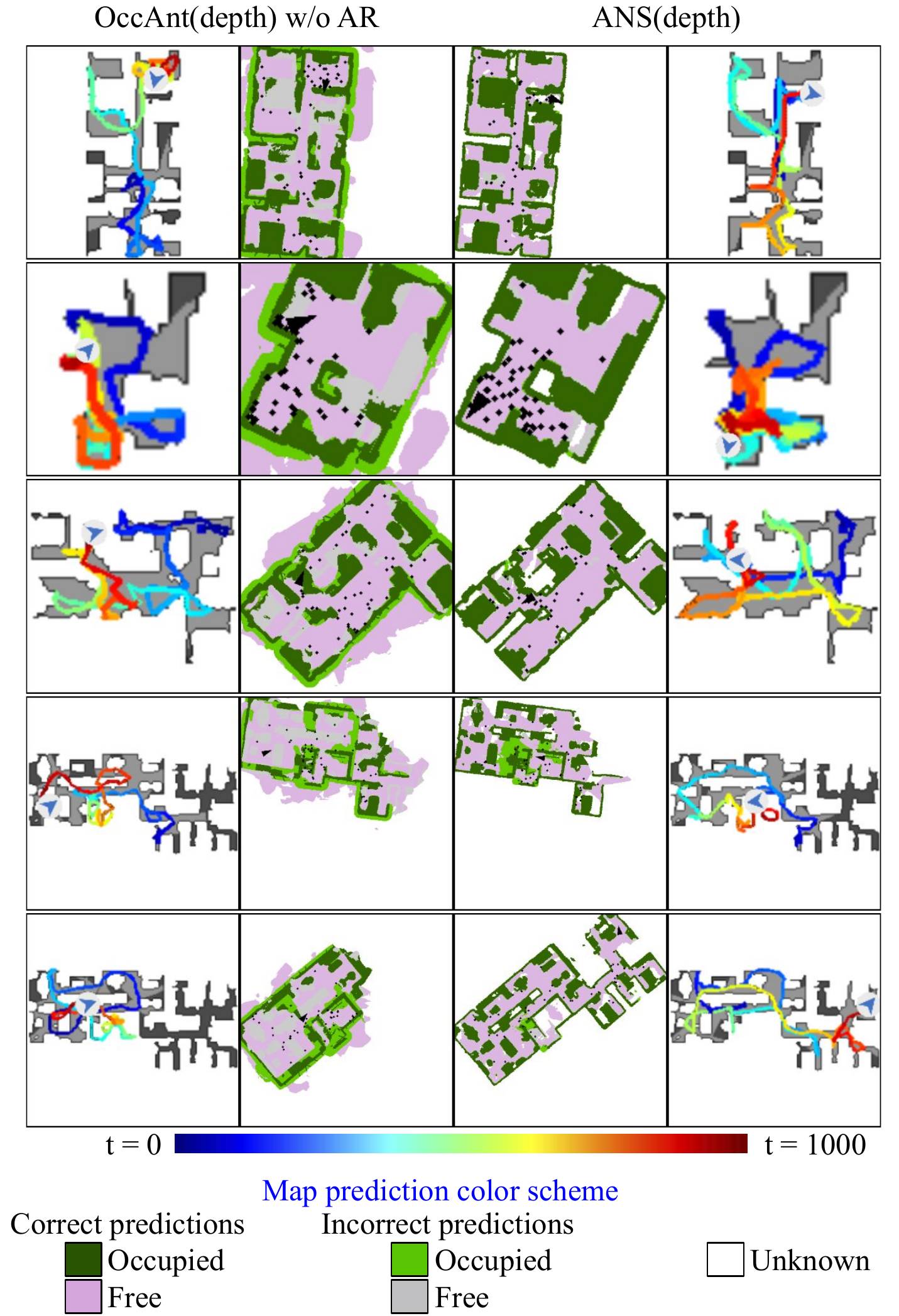}
    \caption{\small We highlight one key weakness of exploration using occupancy anticipation, which is the impact of classification errors in occupancy estimates. We compare OccAnt(depth) w/o AR with ANS(depth) in Gibson. In some cases, OccAnt(depth) w/o AR tends to generate false negatives for occupied regions, classifying some of the explored obstacles as free-space (gray regions in the first 3 rows, 2nd column). While this does not impact the area seen, it does reduce the map quality. On the flip side, OccAnt(depth) w/o AR may prematurely classify some narrow corridors as blocked (similar to Fig.~\ref{fig:occ_ant_negatives}) causing the agent to stop exploring beyond that corridor (light green regions in last two rows, 2nd column). }
    \label{fig:exp_occant_depth_weaknesses}
\end{figure}

\begin{figure}
    \centering
    \includegraphics[width=0.75\textwidth]{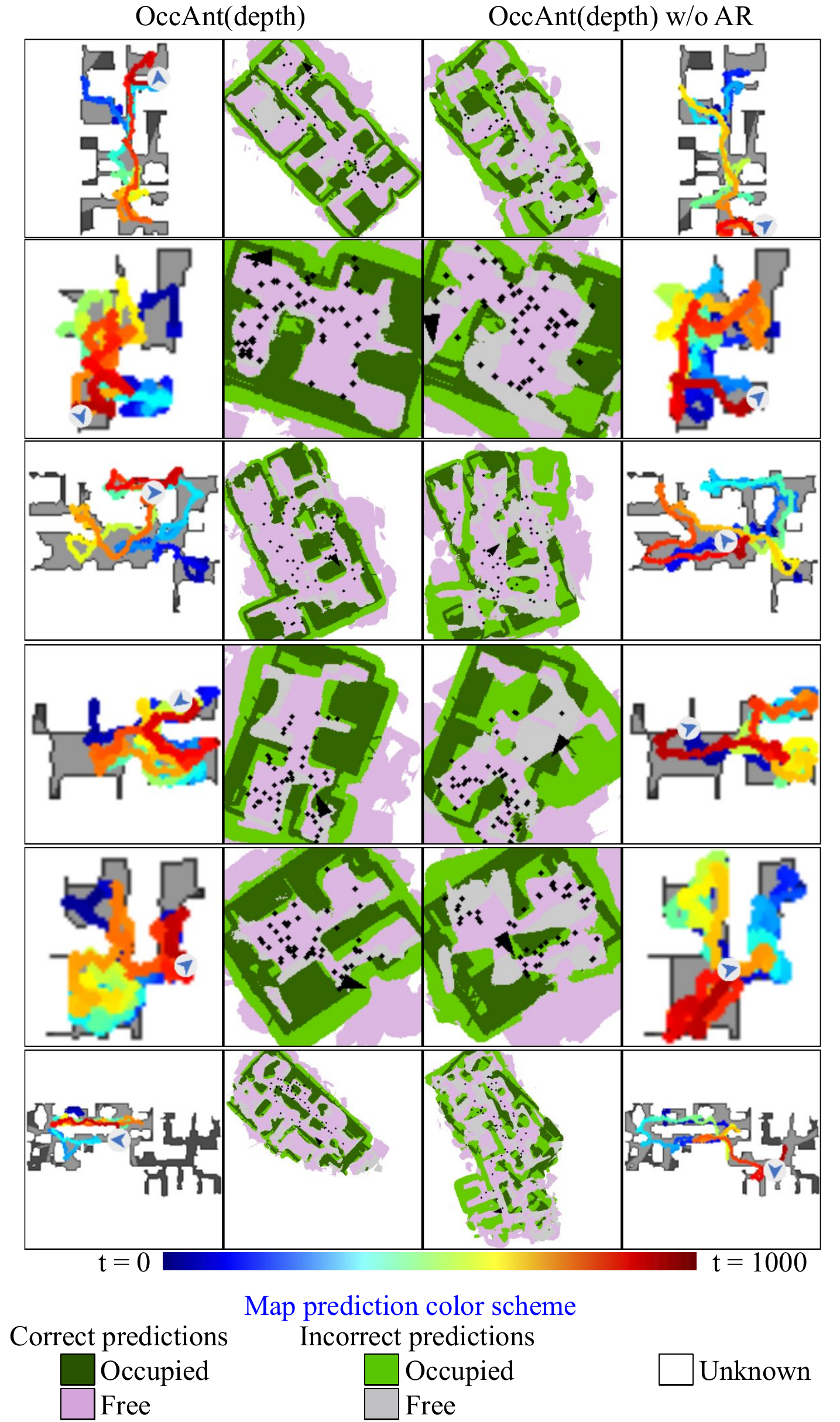}
    \caption{\small \textbf{The impact of using anticipation reward:} In Table~6 from the main paper and Table~\ref{tab:supp_exploration_results} in Supp., we could see that models using anticipation reward generally leads to higher map qualities in noisy test conditions. Here, we show that, when the model that uses the anticipation reward (OccAnt(depth) on the left) accounts much better for the noise in map registration when compared to a vanilla anticipation model that does not use it (OccAnt(depth) w/o AR on the right).}
    \label{fig:exp_occant_anticipation_reward}
\end{figure}

Here, we highlight some example episodes to show that (1) using occupancy anticipation avoids local navigation difficulties and obtains higher map qualities for lower area coverage (Fig.~\ref{fig:exp_occant_depth_strengths}), while sometimes being susceptible to inaccuracies in map predictions (Fig.~\ref{fig:exp_occant_depth_weaknesses}), and (2) the anticipation reward leads to better map registration (i.e., good pose estimates) which results in higher map quality (Fig.~\ref{fig:exp_occant_anticipation_reward}). The  color scheme for the trajectories (from~\cite{habitat19iccv}) and the predicted maps in the center (from~\cite{chaplot2020learning}) are indicated below each plot.

\vfill
\pagebreak

\section{Generating ground-truth for occupancy anticipation}
\label{sec:ground_truth_generation}
\begin{figure}[ht!]
    \centering
    \includegraphics[width=1.0\textwidth]{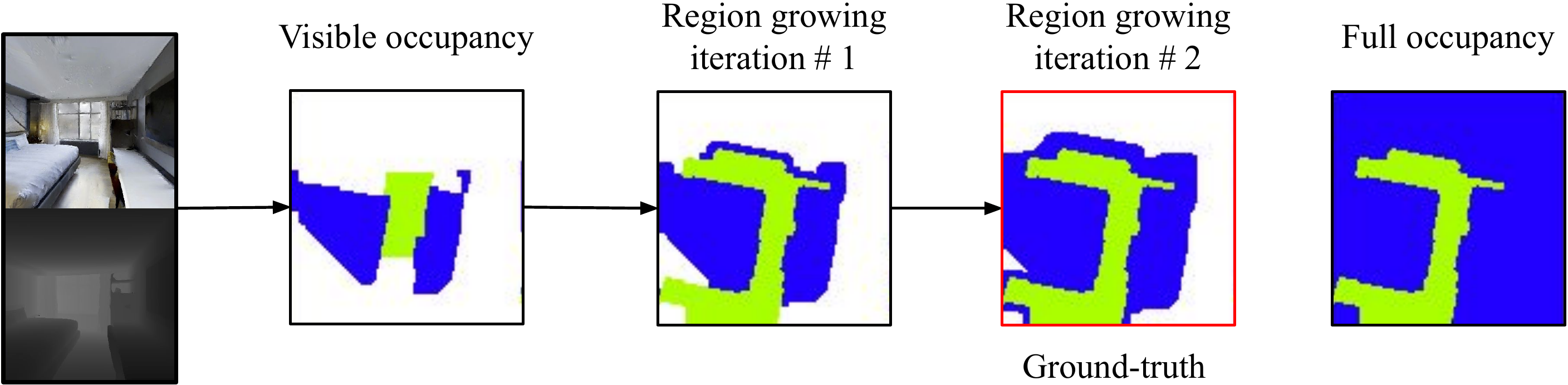}
    \caption{Pipeline for generating anticipation ground-truth.}
    \label{fig:region_growing_figure}
\end{figure}

Using 3D meshes of indoor environments from Gibson and Matterport3D, we obtain the ground-truth local occupancy of a $V \times V$ region in front of the camera which includes parts that may be occluded or outside the field-of-view (see Fig.2 from main paper). However, this may include regions in the environment that are outside the bounds of the environment's mesh. To alleviate this problem, we devise a simple heuristic that generates the ground truth by masking out regions in the occupancy map that are outside the bounds of the environment (highlighted in Fig.~\ref{fig:region_growing_figure}). 

We first obtain the visible occupancy via a geometric projection of the depth inputs (2nd column). We then selectively sample the ground-truth layout (last column) around the visible regions by growing a mask around the visible occupancy by sequential hole-filling and morphological dilation operations. We perform two iterations of this region growing to obtain the final ground-truth used to train our model (3rd \& 4th columns). This heuristic captures the occupied regions that are closer to navigable space in the environment (likely to be objects, walls, etc), while ignoring regions outside the bounds of the environment. This is necessary since the occupancy map from the simulator does not distinguish between obstacles and regions outside the bounds of the environment mesh. Note that these steps apply only in training; during inference the occupancy anticipation proceeds solely in the end-to-end model.

\section{Noise models for actuation and odometry}
\label{sec:noise_models}
Following~\cite{chaplot2020learning}, we simulate realistic actuation and odometry to train and evaluate our exploration agents. For this purpose, we use the PyRobot actuation model provided by Habitat which consists of truncated Gaussians for both the rotation and translation motions.\footnote{\url{https://github.com/facebookresearch/habitat-sim/habitat_sim/agent/controls/pyrobot_noisy_controls.py}} Specifically, we use the default LoCoBot noise-model with the ILQR controller. For simulating noise in the odometry, we similarly use truncated Gaussians for both rotation and translation measurements. For the translation measurement, we use a mean of $0.025\si{m}$ and a standard deviation of $0.001$ For the rotation measurement, we use a mean of $0.9^\circ$ and standard deviation of $0.057^\circ$. The distributions are truncated at 2 standard deviations. These are based on approximate values provided by the authors of ANS.

\section{Differences in ANS implementation}
\label{sec:differences_ans}
We implemented the ANS approach using the published details in \cite{chaplot2020learning} and instructions obtained directly from the authors via private communication as code was not publicly available at the time of our research. Our implementation has a few differences from that in \cite{chaplot2020learning}, which we discuss in the following. For shortest path planning, we use an A* planner instead of fast-marching~\cite{sethian1996fast} used in~\cite{chaplot2020learning} since we were able to find a fast A* implementation that was publicly available.\footnote{A* implementation: \url{https://github.com/hjweide/a-star}}

For aggregating the local occupancy maps $\hat{p}_{t}$ from each observation\footnote{$\hat{p}_{t}$ is the local map at $t$ registered to the global coordinates using the agent's pose estimate.} into the global map $\hat{m}_{t-1}$ from the previous time-step , the authors in~\cite{chaplot2020learning} use channelwise max-pooling of the local and global maps to obtain the updated global map $\hat{m}_{t}$.
\begin{equation}
    \hat{m}_{t} = \text{\code{ChannelwiseMax}}(\hat{m}_{t-1}, \hat{p}_{t})
\end{equation}

Instead, we opt to perform a moving-average over time to allow the agent to account for errors in the map prediction by averaging predictions from multiple views over time.
\begin{equation}
    \hat{m}_{t} = \alpha_{e}\hat{m}_{t-1} +  (1-\alpha_{e})\hat{p}_{t}
\end{equation}
We found that this provided robustness to false positives in the map predictions and registration errors due to odometry noise.

Additionally, since our proposed model anticipates occupancy beyond the visible regions, we found that it is helpful to filter out low-confidence predictions of occupancy on a per-frame basis using the \code{EntropyFilter()} operation.
Given prediction $\hat{p}_t$, \code{EntropyFilter()} masks out the predictions for locations $i,j$ in $\hat{p}_t$ where the binary-entropy of the probabilities across the map channels are larger than a threshold $\tau_{ent}$ before performing the moving-average aggregation.
These low-confidence predictions generally correspond to regions that are hard to anticipate or may have multiple solutions.
Hence, our global map update formula is:
\begin{equation}
    \hat{m}_{t} = \alpha_{e}\hat{m}_{t-1} +  (1-\alpha_{e})\text{\code{EntropyFilter}}(\hat{p}_{t}).
\end{equation}

\vfill
\pagebreak

\section{Implementation details}
\label{sec:implementation_details}
The key hyperparameters for learning the policy and mapper are specified in Table~\ref{tab:hyperparameters}.

\begin{table}[ht!]
\centering
\begin{tabular}{@{}ll@{}}
\toprule
\multicolumn{2}{c}{Policy learning}                             \\ \midrule
Optimizer                          & Adam~\cite{kingma2014adam} \\
$\#$ processes                     & 24                         \\
Learning rate                      & 0.00025                    \\
Value loss coef                    & 0.5                        \\
Entropy coef                       & 0.001                      \\
Discount factor $\gamma$           & 0.99                       \\
GAE $\tau$                         & 0.95                       \\
Episode length                     & 1000                       \\
$\#$ training frames               & 1.5-2 million              \\
PPO clipping                       & 0.2                        \\
PPO epochs                         & 4                          \\
$\#$ PPO minibatches               & 16                         \\
Global policy $\Delta$             & 25                         \\
Global policy update interval      & 20                         \\
Global policy reward scaling       & 0.0001                     \\
Local  policy reward scaling       & 1.0                        \\
Local policy update interval       & 25                         \\
\midrule
\multicolumn{2}{c}{Mapper learning}                             \\ \midrule
Optimizer                          & Adam~\cite{kingma2014adam} \\
Learning rate                      & 0.0001                     \\
Replay buffer size                 & 25000                      \\
Mapper update interval             & 5                          \\
Mapper batch size                  & 32                         \\
Mapper update batches              & 20                         \\
Map scale                          & 0.05$\si{m}$               \\
Local map size (V)                 & 101                        \\
Global map size (G)                & 961                        \\
Aggregation factor ($\alpha_{e}$)  & 0.9                        \\

\bottomrule
\end{tabular}
\caption{Policy and mapper hyperparameters used to train our models}
\label{tab:hyperparameters}
\end{table}

\section{Occupancy anticipation architecture}
\label{sec:occupancy_anticipation_architecture}
\begin{figure}[ht!]
    \centering
    \includegraphics[width=\textwidth]{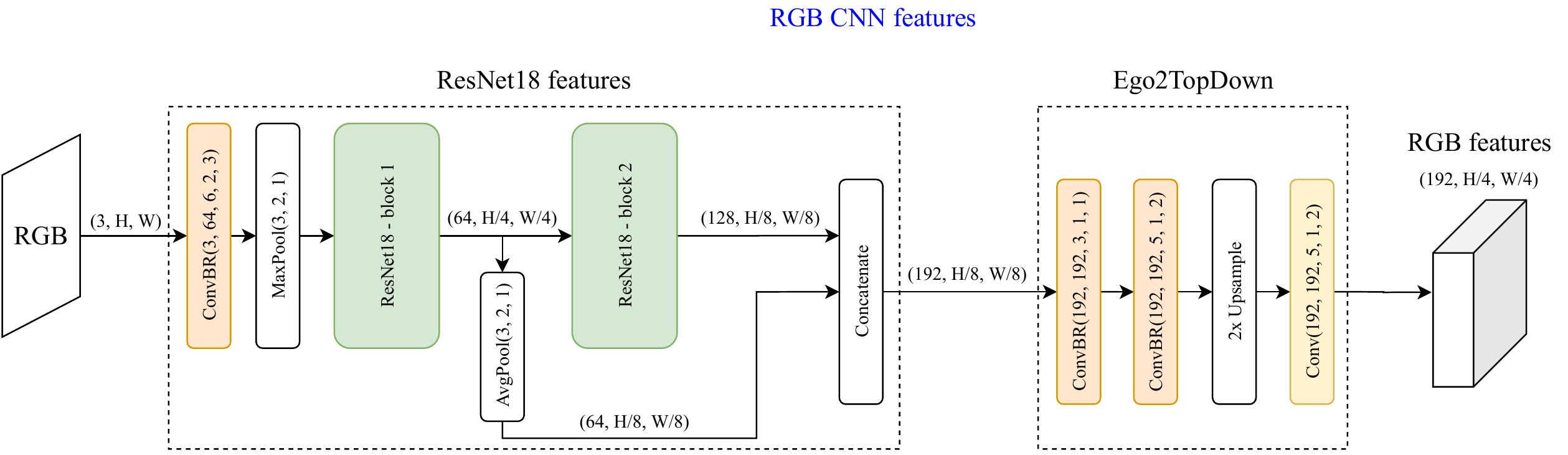}
    \caption{\small\textbf{RGB CNN features:} extracts features from RGB images using ResNet18 blocks, and further processes the features to obtain compatible RGB features in a top-down view.}
    \label{fig:rgb_cnn_features}
\end{figure}

\begin{figure}[ht!]
    \centering
    \includegraphics[width=\textwidth]{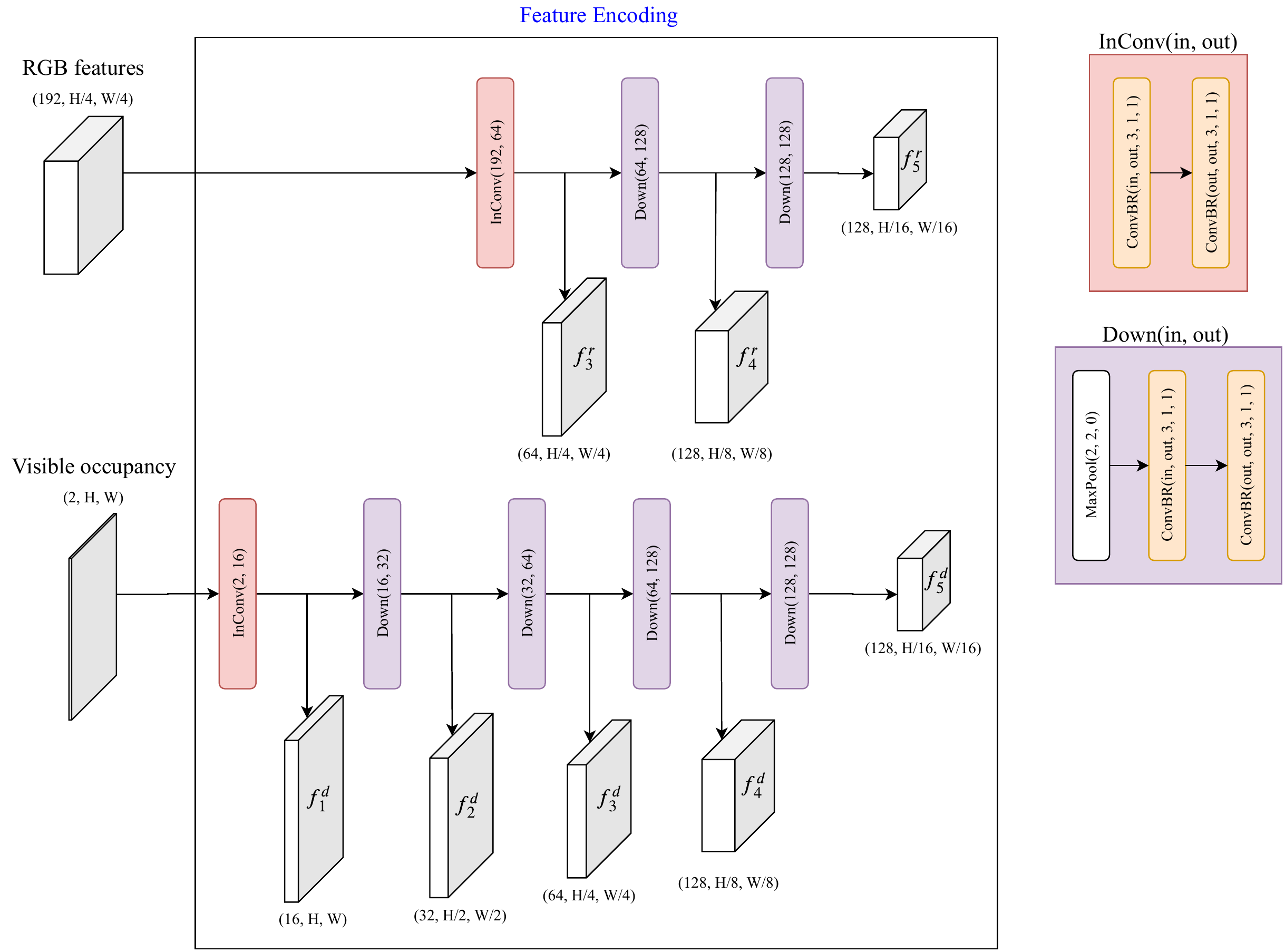}
    \caption{\small\textbf{Feature encoding:} The RGB features and visible occupancy are encoded using independent UNet encoding layers. The expanded view of the ``InConv" and ``Down" blocks are shown on the right. The encoded RGB and visible occupancy features at different levels are $f_{3:5}^{r}$ and $f_{1:5}^{d}$, respectively.}
    \label{fig:feature_encoding}
\end{figure}

\begin{figure}[ht!]
    \centering
    \includegraphics[width=\textwidth]{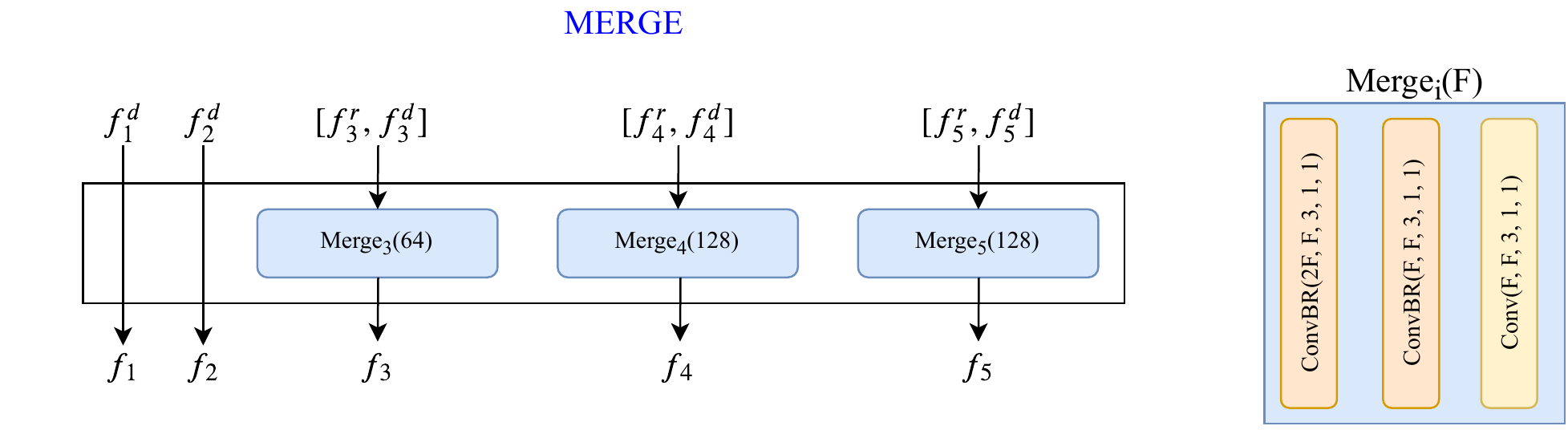}
    \caption{\small \textbf{MERGE:} combines the RGB ($f_{3:5}^{r}$) and visible occupancy ($f{1:5}^{d}$) features obtained from Feature encoding layers in a layerwise fashion to obtain a set of merged features $f = f_{1:5}$. Since the RGB features are not available at levels $1$ and $2$, it simply uses the visible occupancy features for those levels. The expanded view of the ``$\text{Merge}_{i}$(F)" block is shown on the right.}
    \label{fig:merge}
\end{figure}

\begin{figure}[ht!]
    \centering
    \includegraphics[width=\textwidth]{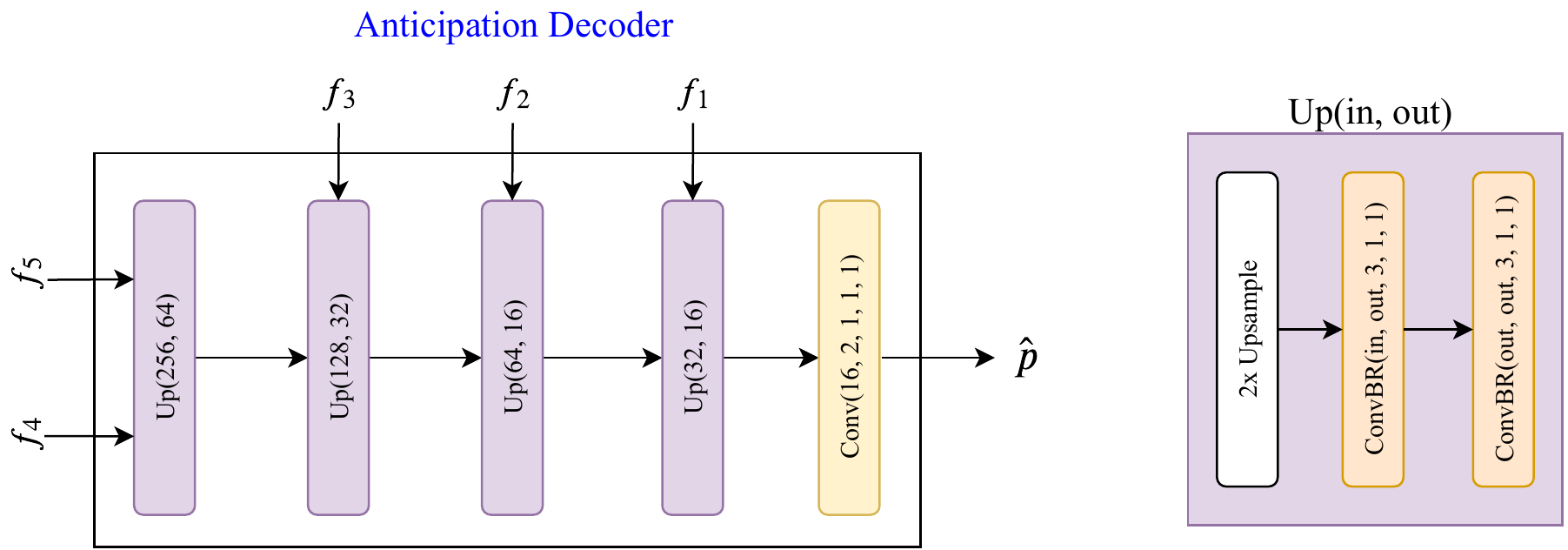}
    \caption{\small \textbf{Anticipation Decoder:} a typical UNet decoder that takes features provided by MERGE ($f = f_{1:5}$) and decodes them using residual connections to obtain the anticipated occupancy map $\hat{p}$. The expanded view of the ``Up" block is shown on the right. }
    \label{fig:anticipation_decoder}
\end{figure}

The architecture diagrams for the individual components of our occupancy anticipation model (Fig. 2 in main paper) are shown in Figs.~\ref{fig:rgb_cnn_features},~\ref{fig:feature_encoding},~\ref{fig:merge} and~\ref{fig:anticipation_decoder} with a brief description of the role of each module. We follow the PyTorch~\cite{NEURIPS2019_9015} conventions to describe individual layers, with the tensor shapes represented in (C, H, W) notations. The descriptions for individual layers are:

\begin{itemize}
    \item \textbf{ConvBR:} a combination of \code{nn.Conv2d}, \code{nn.BatchNorm2d} and \code{nn.ReLU} layers with the arguments representing the input channels, output channels, kernel size, stride and padding.
    \item \textbf{MaxPool:} an instantiation of the \code{nn.MaxPool2d} layer with the arguments representing the kernel size, stride and padding.
    \item \textbf{Conv:} a \code{nn.Conv2d} layer with the arguments representing the input channels, output channels, kernel size, stride and padding.
    \item \textbf{AvgPool:} an instantiation of the \code{nn.AvgPool2d} layer with the arguments representing the kernel size, stride and padding.
    \item \textbf{2x Upsample:} an instantiation of the \code{nn.Upsample} layer with a scaling factor of $2$.
\end{itemize}

\section{ANS projection unit architecture}
\begin{figure}[ht!]
    \centering
    \includegraphics[width=\textwidth]{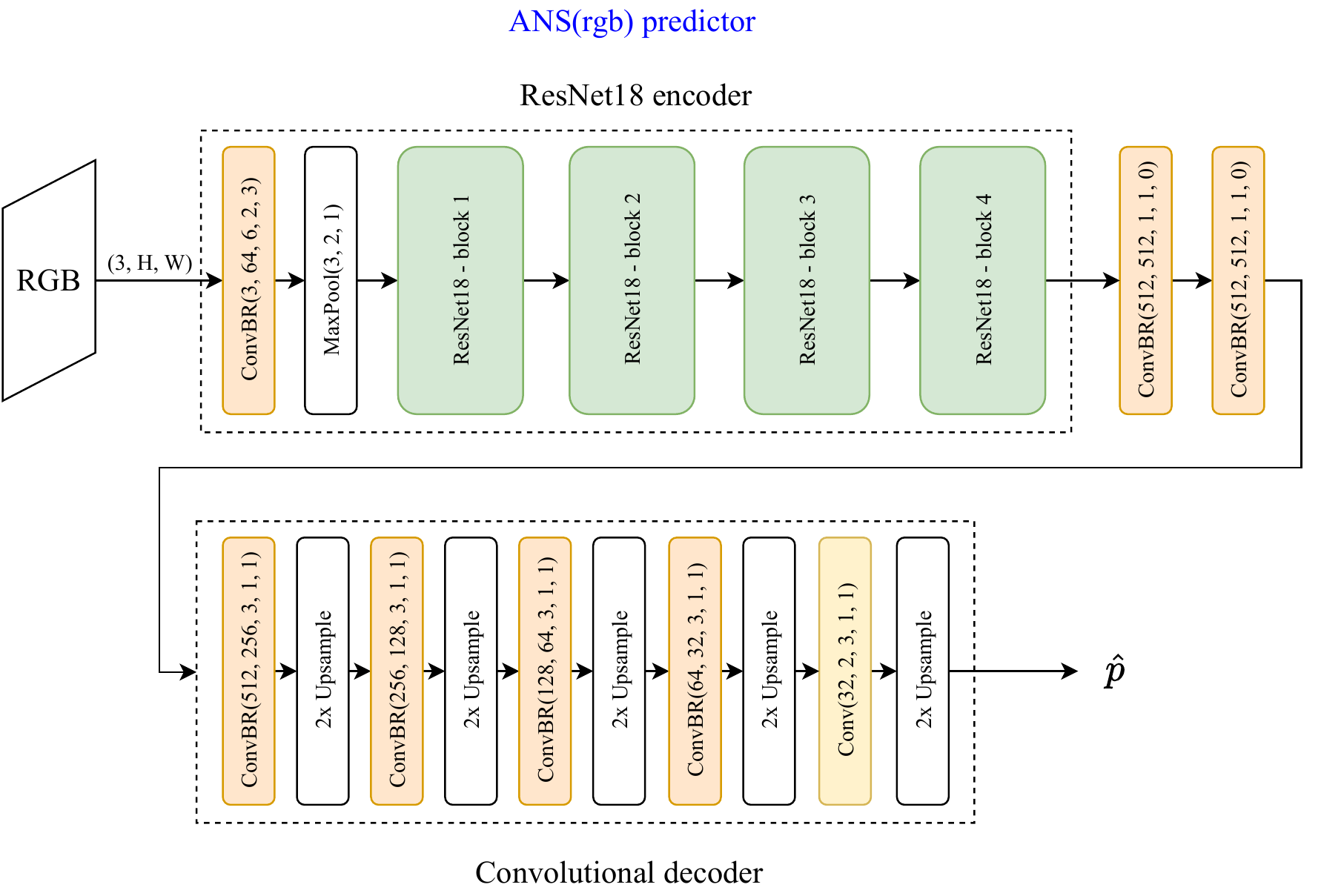}
    \caption{\small \textbf{ANS projection unit:} ResNet-18 features are extracted, followed by two fully connected layers (represented by $1\times 1$ convolutions) and a convolutional decoder that uses ``Upsample" blocks to increase the output resolution and predict the occupancy estimates $\hat{p}$. Note that this is supervised to predict the visible occupancy map, not the anticipated occupancy map (see Fig. 1 in main paper).}
    \label{fig:ans_projection_unit}
\end{figure}

The projection unit architecture for the ANS(rgb) baseline is shown in Fig.~\ref{fig:ans_projection_unit}. This is based on the architecture in~\cite{chaplot2020learning} with some minor differences. It uses \code{nn.BatchNorm} + \code{nn.ReLU} blocks instead of \code{nn.Dropout} in the fully connected layers, it has a larger convolutional decoder to account for our larger map outputs, and it consists of \code{nn.Conv2d} + \code{nn.Upsample} layers instead of than \code{nn.ConvTranspose2D} layers as this has been shown to reduce checkerboard artifacts~\cite{odena2016deconvolution}.

\section{View extrapolation baseline}
\label{sec:view_extrapolation_baseline}
\begin{figure}[ht!]
    \centering
    \includegraphics[width=\textwidth]{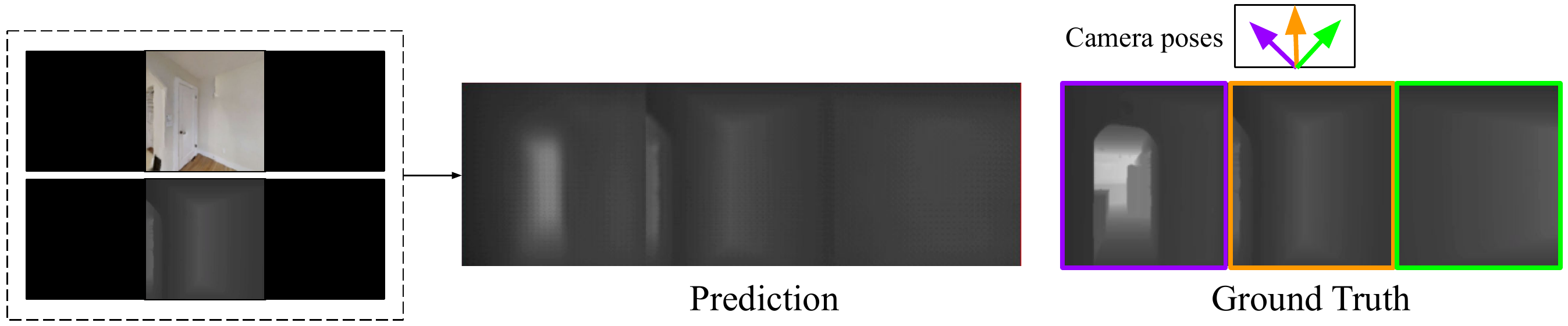}
    \caption{\small \textbf{View extrapolation task:} Given the agent's egocentric RGB-D input, we predict the depth-map for two additional depth-sensors placed at $45^{\circ}$ angles to the left (purple) and right (green) of the central input (orange). These are geometrically projected to the top-down view to obtain the occupancy estimates.}
    \label{fig:view_extrapolation_task}
\end{figure}

\begin{figure}[ht!]
    \centering
    \includegraphics[width=\textwidth]{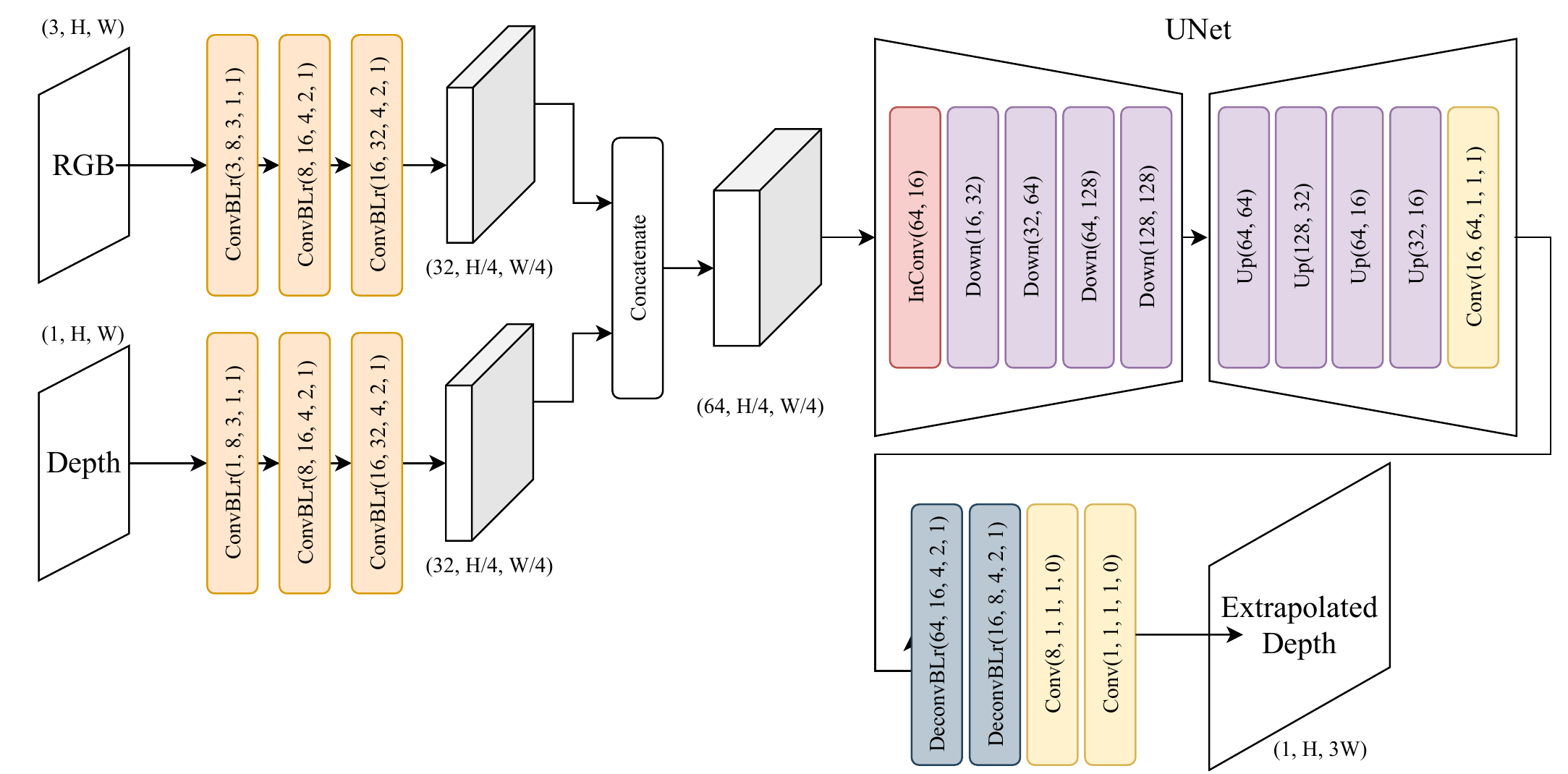}
    \caption{\small \textbf{View extrapolation architecture~\cite{Yang_2019_CVPR}:} The $90^\circ$ FoV RGB and depth inputs are independently encoded using Convolutional layers, concatenated and processed using a UNet model. The decoded features from UNet are used to extrapolate the final depth predictions. ``DeconvBLr" uses \code{nn.ConvTranspose2D} to perform the upsampling. Note that ``ConvBLr" and ``DeconvBLr" use \code{nn.LeakyReLU(0.1)} instead of ``nn.ReLU()". }
    \label{fig:view_extrapolation_architecture}
\end{figure}

We now provide more details on the task-defintion and architecture for the View-extrap. baseline introduced in Sec. 4.1 in the main paper. The goal is to extrapolate $180^\circ$ FoV depth from $90^\circ$ FoV RGB-D inputs in order to evaluate the performance of scene completion approaches~\cite{song2018im2pano3d,Yang_2019_CVPR}. Since Habitat~\cite{habitat19iccv} does not natively support panoramic rendering, we use a simpler solution to account for this. We place two cameras with $\pm45^\circ$ heading offsets and aim to regress those from the egocentric view (see Fig.~\ref{fig:view_extrapolation_task}). Since each camera has a $90^{\circ}$ FoV, this leads to an effective coverage of $180^{\circ}$ once the agent anticipates the unobserved portions. We base our architecture for view extrapolation on the model from~\cite{Yang_2019_CVPR} with a capacity similar to our model to permit online training during policy learning (see Fig.~\ref{fig:view_extrapolation_architecture}). It takes as input the $90^\circ$ FoV RGB-D images and regresses the left and right cameras. It is trained to minimize the pixelwise $\ell_{1}$ loss between the prediction and the ground-truth.

\section{Comparing the model capacities of different methods}
\label{sec:model_capacity_comparison}

We compare the overall model capacity of our approaches with the baselines in Table~\ref{tab:model_capacity_comparison}. The depth-only models (bottom 2 rows) tend to have fewer parameters than the rgb-only models as they rely on geometric projection for processing depth (no ResNet backbone). Our depth model has comparable number of parameters with the depth-only baselines. Our rgb model has slightly more parameters than the rgb baseline. However, this is due to the fact that OccAnt(rgb) takes the output of ANS(rgb) as an additional input. However, since ANS(rgb) is kept frozen throughout the training of OccAnt(rgb), this effectively gives us 5.7M trainable parameters. 

\begin{table}[!]
    \centering
    \begin{tabular}{@{}lc@{}}
    \toprule
    Method        & Parameters (in millions) \\ \midrule
    ANS(rgb)      & 14.16                   \\
    OccAnt(rgb)   & 19.86                   \\ \midrule
    ANS(depth)    & 0.87                    \\
    OccAnt(depth) & 1.72                    \\\bottomrule
    \end{tabular}
    \caption{Comparing model capacity of different approaches}
    \label{tab:model_capacity_comparison}
\end{table}

\section{Habitat Challenge 2020}
\label{sec:habitat_challenge_2020}
We detail the key issues we had to address for the PointNav track of Habitat Challenge 2020~\cite{habitat-challenge} and the changes to our system required to achieve our state-of-the-art results. Compared to the 2019 Habitat Challenge, there were two key changes that increased the task difficulty:

\paragraph{Lack of GPS+Compass sensor:}
The presence of the GPS+Compass sensor used in earlier challenges continually provides the agent with a perfect estimate of the position and heading angle of the goal relative to its current position. Such perfect localization has been exploited in the past by purely geometric~\cite{gupta2017cognitive} and learned~\cite{wijmans2019decentralized} approaches to achieve high-quality PointNav performance. However, such high precision localization is hard to achieve in practice. The 2020 challenge instead requires navigation in the absence of the GPS+Compass sensor. Instead, the goal location is only specified initially at the start of the episode, requiring the agent to accurately keep track of its position in the environment to successfully reach the goal.

\paragraph{Noisy actuation and sensing:} In the 2020 challenge, RGB-D sensing noise is simulated artificially by using a Gaussian noise-model for the RGB sensor and the Redwood noise model~\cite{Choi_2015_CVPR} for the depth sensor. Additionally, actuation noise in the robot motions is simulated by using a noise model obtained from the LoCoBot~\cite{pyrobot2019}.

We adapted our model in several ways to address these challenges. To address the lack of GPS+Compass sensor, we used an online pose estimator that uses RGB-D inputs $x_{t}$ and $x_{t+1}$ to estimate the relative change in the pose $\Delta p_{t+1}$. These pose changes are summed up over time to track the agent's pose $p_{t+1}$. When compared to the original ANS model, we found that using RGB-D inputs gave slightly better estimates and was more computationally efficient than using top-down maps. The pose estimator consists of a 6 convolutional layers followed by 3 fully-connected layers to predict the pose for each modality (RGB, depth) independently. The predictions are combined by using input-conditioned weighting factors that are estimated using a learned MLP with 4 fully-connected layers.

\begin{figure}
    \centering
    \includegraphics[width=0.9\textwidth]{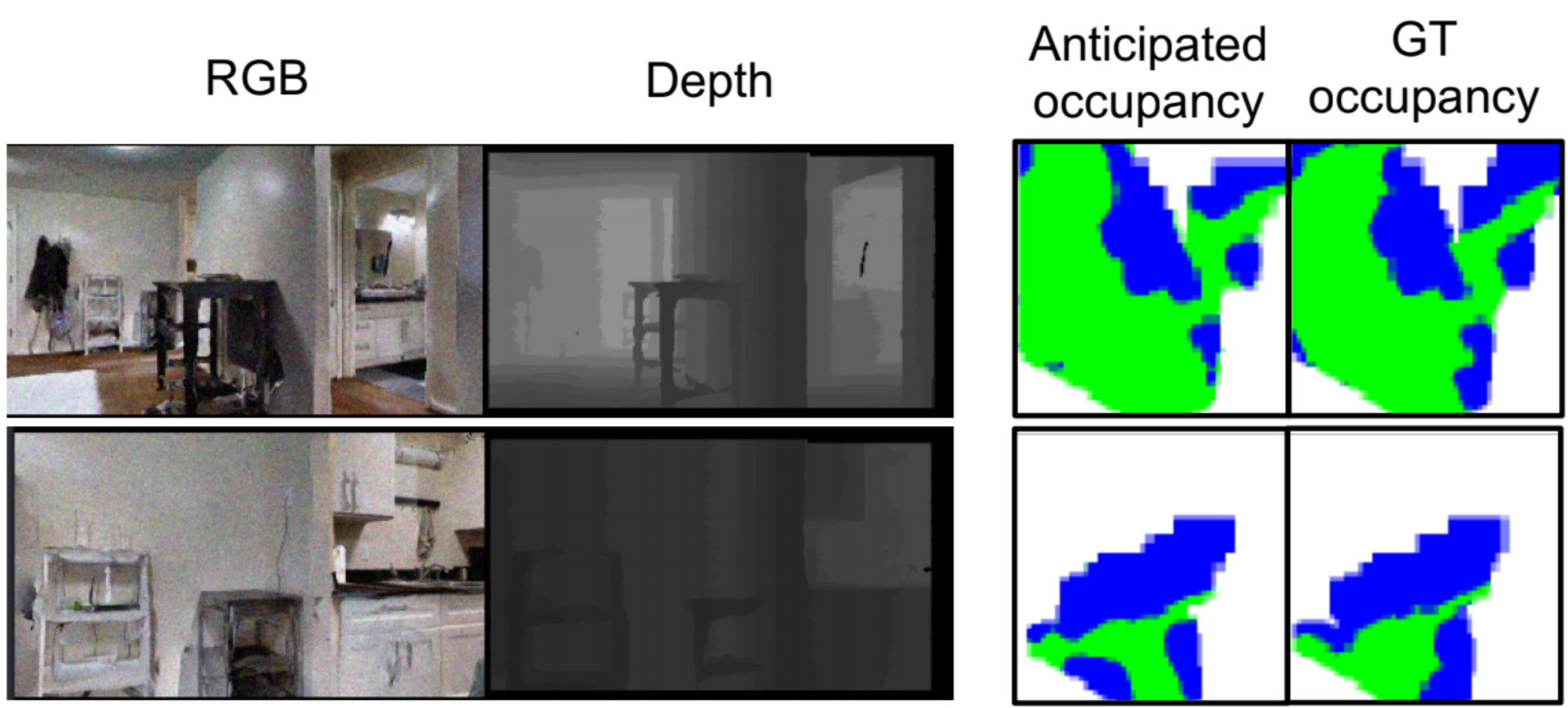}
    \caption{Qualitative results from the 2020 Habitat Challenge: On the left, we show the noisy RGB and depth inputs. On the right, we show the corresponding anticipated and ground-truth occupancy. Our model learns to anticipate accurately in the presence of noise.}
    \label{fig:challenge_qualitative}
\end{figure}

To handle noisy sensing, we train our occupancy anticipation model end-to-end on the noisy inputs, which gave accurate predictions (see Fig.~\ref{fig:challenge_qualitative}). We found that OccAnt(depth) gave the best performance, and that adding RGB information to occupancy anticipation did not lead to significant changes in performance. 

To deal with noisy actuation, we found that the learned pose estimator gave robust estimates of the agent position. Despite having this pose estimator, we experienced large drifts in the estimate over time due to high variance in the actuation noise. To partially mitigate this issue, we focused on efficient navigation with safe planning that maintains sufficient distance from obstacles while planning shortest paths. In practice, we found that reducing the number of collisions leads to faster navigation and lower drift in the pose estimates. We achieve this by using a weighted variant of the classic A-star search algorithm~\cite{Hart1968}.\footnote{Weighted A-star implementation: \url{https://github.com/srama2512/astar_pycpp}}

Additionally, we incorporated some simple heuristics from the original Active Neural SLAM implementation to update the occupancy maps based on collisions, and used an analytical local policy for navigation instead of a learned policy.

\end{document}